\documentclass[letterpaper, 10 pt, conference]{ieeeconf}
\IEEEoverridecommandlockouts

\usepackage[utf8]{inputenc}
\usepackage{float}
\usepackage{cite}
\usepackage{amsmath,amssymb,amsfonts}
\usepackage{hyperref}
\usepackage{graphicx}
\usepackage{textcomp}
\usepackage{xcolor}
\usepackage{caption}
\usepackage{subfigure}
\usepackage{subcaption}
\usepackage{tabularx}
\usepackage{algorithm}
\usepackage{algpseudocode}
\usepackage{cleveref}
\graphicspath{{Figures/}}
\usepackage{mdframed}
\usepackage{array}
\usepackage[table,dvipsnames]{xcolor}
\newcolumntype{C}{>{\centering\arraybackslash}p{2cm}}
\usepackage{tikz}
\usepackage[intoc]{nomencl}
\makenomenclature

\newcommand{\M}{\mathcal{M}}
\newcommand{\p}{\textbf{p}}
\newcommand{\q}{\textbf{q}}
\newcommand{\s}{\textbf{r}}
\newcommand{\h}{\textbf{h}}
\newcommand{\id}{\mathbf{i}}

\newcommand{\X}{\textit{X}}
\newcommand{\tm}{\textit{t}}

\newcommand{\calg}[1]{\mathcal{#1}}

\DeclareMathOperator*{\argmin}{argmin}

\title{\LARGE\bf Where to Fly, What to Send: Communication-Aware Aerial Support for Ground Robots}

\author{
Harshil Suthar and Dipankar Maity
\thanks{
The authors are with the Department of Electrical and Computer Engineering at the University of North Carolina at Charlotte, NC, USA, 28223. (e-mails: {\tt \{hsuthar1, dmaity\}@charlotte.edu}).
}
}

\begin{document}

\newpage

\maketitle

\begin{abstract}
In this work we consider a multi-robot team operating in an unknown environment where one aerial agent is tasked to map the environment and transmit (a portion of) the mapped environment to a group of ground agents that are trying to reach their goals.
The entire operation takes place over a bandwidth-limited communication channel, which motivates  the problem of determining \textit{what} and \textit{how much} information the assisting agent should transmit and \textit{when} while simultaneously performing exploration/mapping.   
The proposed framework enables the assisting aerial agent to decide what information to transmit based on the \textit{Value-of-Information} (VoI), how much to transmit using a \textit{Mixed-Integer Linear Programming} (MILP), and how to acquire additional information through an utility score-based environment exploration strategy.  
We perform a communication-motion trade-off analysis between the total amount of map data communicated by the aerial agent and the navigation cost incurred by the ground agents.
\end{abstract}

\section{Introduction} \label{INTRODUCTION}
Recent advances in distributed control, communication infrastructure, and reinforcement learning have accelerated the development of multi-agent systems. 
Heterogeneous robot teams now find applications across diverse domains, including search and rescue in unknown environments~\cite{shen2017collaborative}, precision agriculture~\cite{tokekar2016sensor}, warehouse inspection~\cite{ribeiro2022collaborative}, and planetary exploration~\cite{schuster2020arches}.  
In these tasks, agents coordinate, cooperate, and collaborate to enhance overall performance through effective work distribution, spatial coverage, and specialization~\cite{prorok2021beyond}.
Multi-agent systems form a network of autonomous agents that exchange information and process data independently, thereby facing challenges related to computation, communication, and storage~\cite{nowzari2019event}.

Several studies have investigated the collaboration between autonomous aerial vehicles (UAVs) and autonomous ground vehicles (UGVs), where agents complement each other’s distinct capabilities. 
In such heterogeneous teams, UAVs explore the environment, collect remote observations, and share map or path information with UGVs to support their navigation.
However, information exchange in such systems can be affected by factors such as occlusion, range limitations, noise, and bandwidth constraints.
The problem becomes more challenging when a UAV must support multiple UGVs under limited bandwidth, requiring decisions on what information to transmit to whom, how much to share, and which regions to explore to gather informative data.
To ensure the team’s operational efficiency and overall performance, these communication constraints must be accounted for during the planning and decision-making process~\cite{gielis2022critical}.

Building on the discussed context and motivation, this work aims to develop and evaluate a collaborative framework for a team comprising a single UAV and multiple UGVs operating in an unknown environment with limited communication and no prior knowledge.
The proposed framework enables agents to selectively share task-relevant information, allocate bandwidth efficiently among collaborators, and adopt effective exploration strategies to enhance overall team performance.

\begin{figure}
    \centering
    \includegraphics[width = \linewidth]{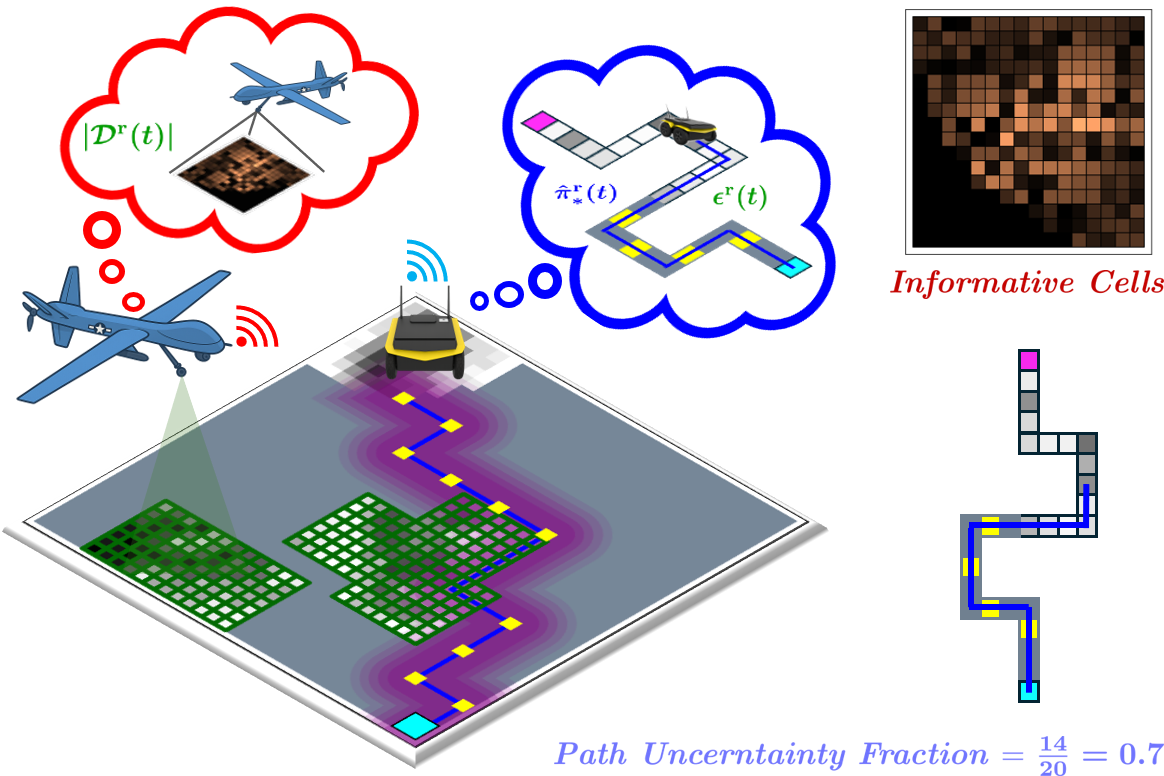}
\caption{\small Overview of a helper/UAV assisting the $\mathbf{r}^{th}$ receiver/UGV agent by sending remote map observations based on the receiver's need. The UAV selects cells to send from the set of informative cells. Brighter the cell color the higher its \textit{value-of-information}.  UGV sends its estimated path way-points shown by yellow cells and \textit{path-uncertainty} fraction.}
\label{Figure: 1} 
\vspace{-4mm}
\end{figure}

\subsubsection*{\textbf{Related Works}}  
Map-compression-based approaches improve inter-agent communication efficiency. Corah et al.~\cite{8633953} proposed a framework where a multi-robot team explores a 3D environment using a shared global map.
Each robot compresses its local point-cloud into a Gaussian Mixture Model (GMM), representing occupied regions with ellipsoids.
Agents have to reconstruct detailed maps from the GMM for planning and use a finite-horizon Monte Carlo tree-search planner for exploration.
In~\cite{9140424}, the authors proposed an information-theoretic map compression framework that generates an abstracted map based on the agent’s available computing resources. 
Using hierarchical data structures such as quad-trees or oct-trees, the method produces multi-resolution cells, assigning finer resolution to high-information regions and compressing areas with low task relevance.
Psomiadis et al.~\cite{psomiadis2023communicationawaremapcompressiononline} extended this idea with a communication-aware framework in which the aerial agent optimally compresses its local map using path information from the ground agent. In~\cite{psomiadis2025communicationawareiterativemapcompression}, the approach is further extended by introducing a decoder capable of iterative map estimation, handling noise through Kalman filter techniques, though the aerial agent’s path in both remains predefined.
However, these methods are limited to a single-UAV, single-UGV setup and do not address what information should be sent to whom, how bandwidth should be allocated, or how new data should be gathered so that all agents can benefit from exploration.

Learning-based methods have also been found to be effective in improving communication within multi-agent systems by enabling agents to learn communication schedules and determine what information to share.
Liu et al.~\cite{9156848} proposed a framework that learns when and with whom to communicate, reducing bandwidth during inference. The extension in~\cite{9197364} adds a three-stage handshake to match available and requested perception data. Yue et al.~\cite{hu2022where2commcommunicationefficientcollaborativeperception} used spatial confidence maps to create compact, task-relevant messages.
Kim et al.~\cite{kim2019learningschedulecommunicationmultiagent} introduced \texttt{SchedNet}, which schedules agents with high-value observations under bandwidth constraints. Li et al.~\cite{li2024contextawarecommunicationmultiagentreinforcement} developed a context-aware communication protocol where agents first share short context messages and then exchange personalized responses via attention mechanisms.
While these learning-based methods improve communication efficiency, they require extensive training, generalize poorly, and impose high computational costs on resource-limited~agents.

\subsubsection*{\textbf{Contribution}}  
The main contributions of this work are: 
\begin{enumerate}
    \item A strategic, utility-based exploration approach for gathering additional information to assist a team of agents.  
    \item A communication protocol based on a \textit{Value-of-Information} (VoI) principle, enabling the UAV to select relevant requested information for each UGV agent.  
    \item A task-aware bandwidth allocation strategy formulated using the \textit{Mixed-Integer-Linear-Programming} (MILP).  
    \item Simulation-based experimental results and evaluation of the proposed framework.  
\end{enumerate}

\subsubsection*{\textbf{Paper Organization}}  
The rest of the paper is organized as follows: 
\Cref{PRELIMINARIES} provides background information and presents the problem setup and formulation. 
\Cref{Proposed Framework} describes the components of the proposed framework. 
In~\Cref{Simulation Results}, the simulation setup and performance analysis are discussed. 
Finally, the paper is concluded in~\Cref{Conclusion}.


\section{Relevant background and problem setup} \label{PRELIMINARIES}

\subsection{Preliminaries: Data Transfer Map and Information Map}\label{subsec:prelim}

We represent the environment as a 2D \footnote{For a 3D environment, a 3D occupancy grid is considered.} 
occupancy grid map $\mathcal{M}$ with dimensions $N \times N$, where $N \in \mathbb{Z}^{++}$.
Each cell in the grid is identified by its center coordinates $(x, y)$ in the grid frame, where $x, y \in [1, N]$.
Let $\mathbf{p}$ denote a generic cell location, i.e., $\exists (x, y)$ such that $\mathbf{p} = (x, y)$.
The occupancy value of a cell is denoted by $o_{\p}$.
All grid cell locations $\p \in \M$ are considered traversable if $o_{\p} \leq \phi^{\text{obs}}$, where $\phi^{\text{obs}} \in [\phi^{\min}, \phi^{\max}]$ denotes the occupancy threshold separating traversable and obstacle regions; $\phi^{\min}$ and $\phi^{\max}$ denote the minimum and maximum occupancy values, respectively.
The sets of explored and unexplored cells in the environment map are denoted by $\M_{e}$ and $\M_{u}$, respectively.
Each unexplored cell $\p \in \mathcal{M}_{u}$ is assigned an occupancy value $\phi^{u} (\le \phi^{\text{obs}})$ for planning purposes, i.e., each unknown cell is optimistically assumed to be traversable. This optimistic assumption ensures completeness of the proposed algorithm (i.e., a path will be found if such exists).

\begin{figure}
    \centering
    \subfigure[]{
    \includegraphics[width=0.23\linewidth]{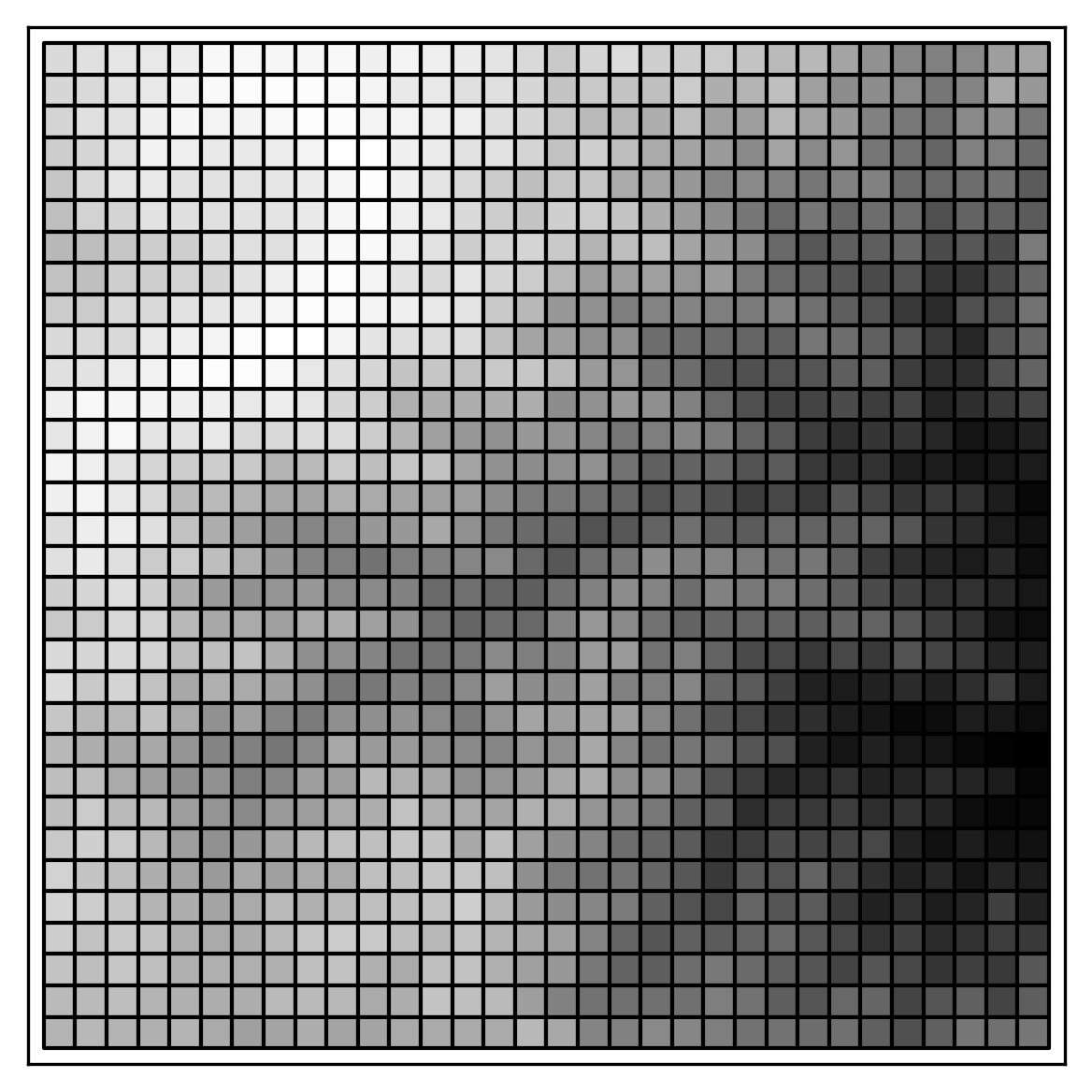}
    \label{Fig:1a}
    } \hspace{ -5mm}
    \subfigure[]{
    \includegraphics[width=0.23\linewidth]{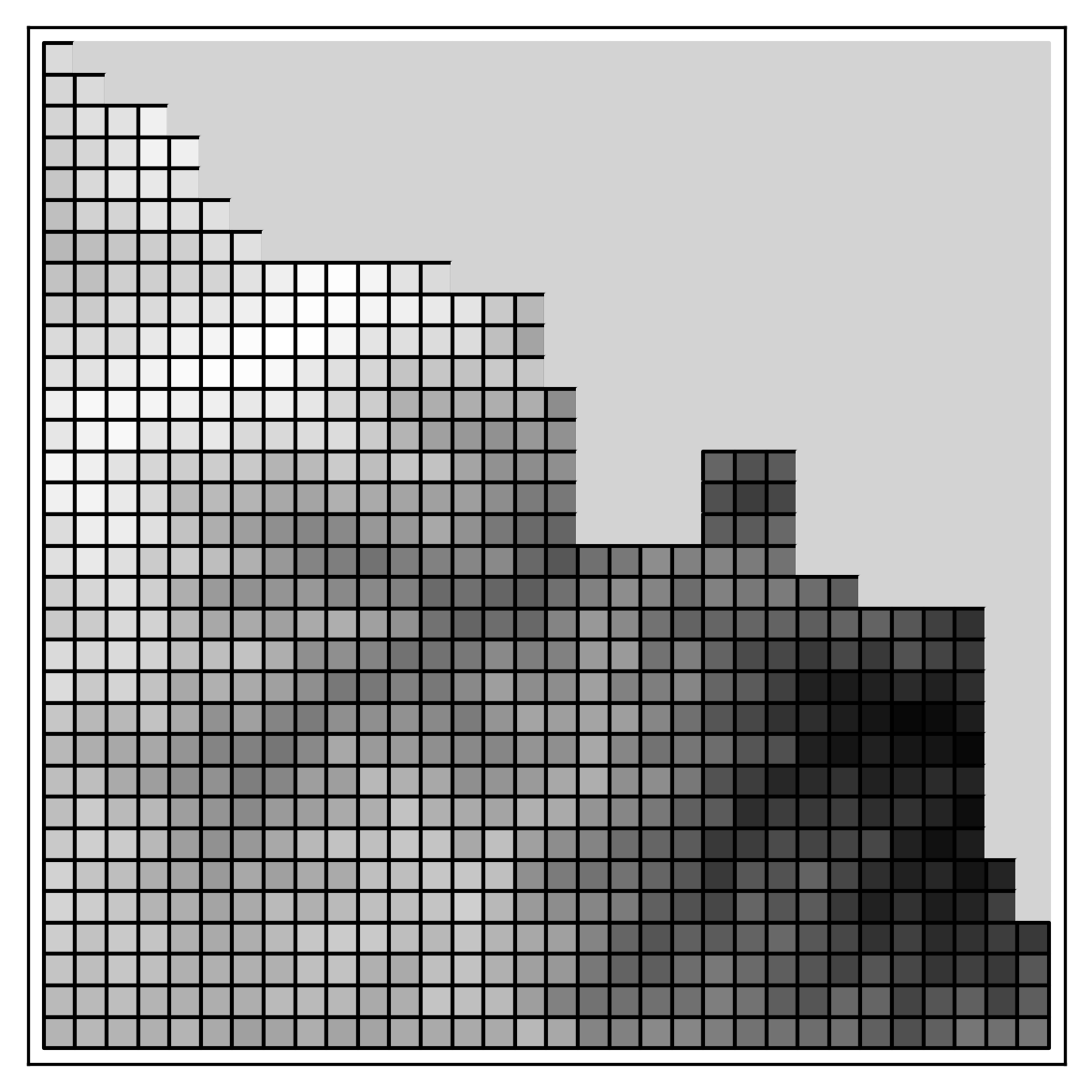}
    \label{Fig:1b}
    } \hspace{ -5mm}
    \subfigure[]{
    \includegraphics[width=0.23\linewidth]{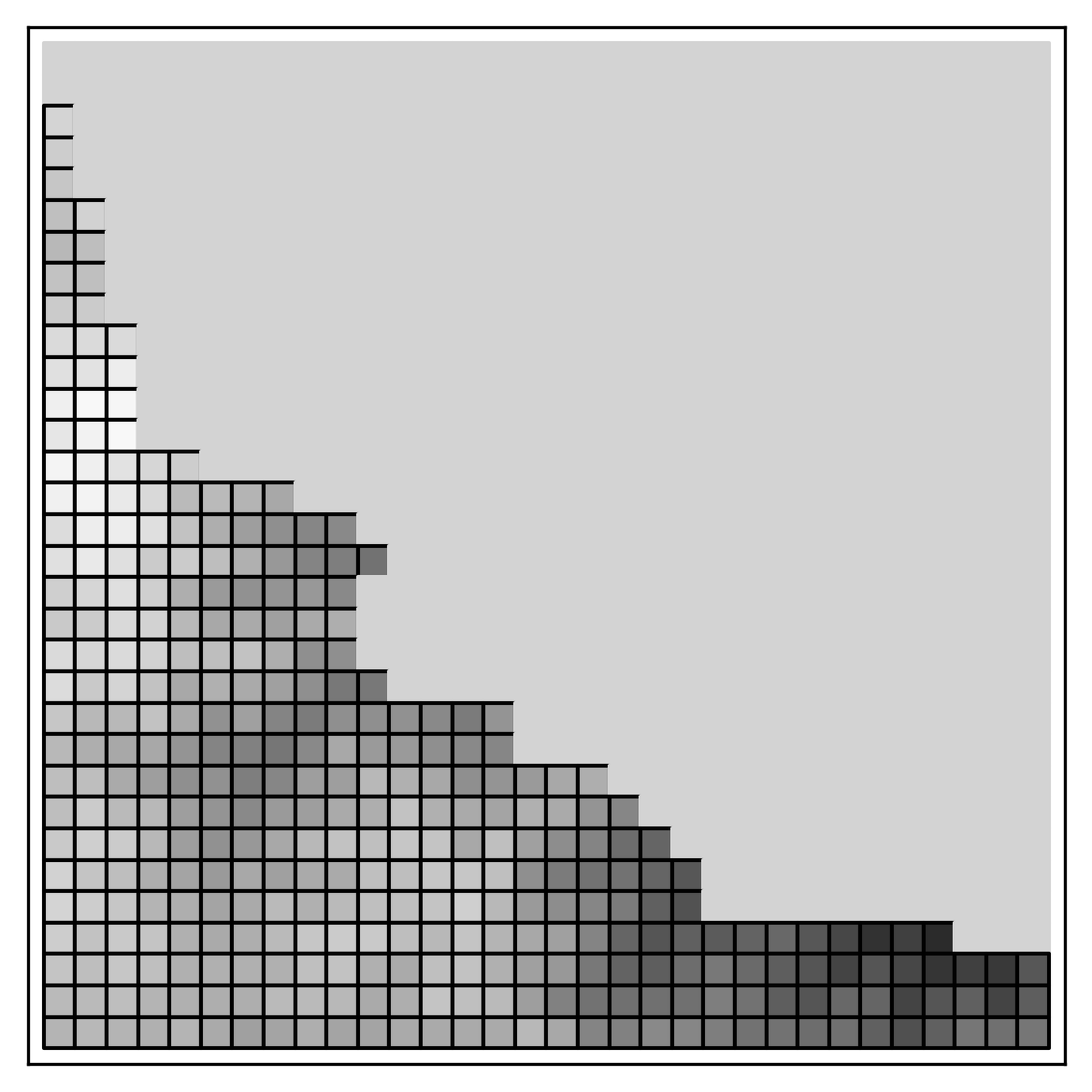}
    \label{Fig:1c}
    } \hspace{ -5mm}
    \subfigure[]{
    \includegraphics[width=0.23\linewidth]{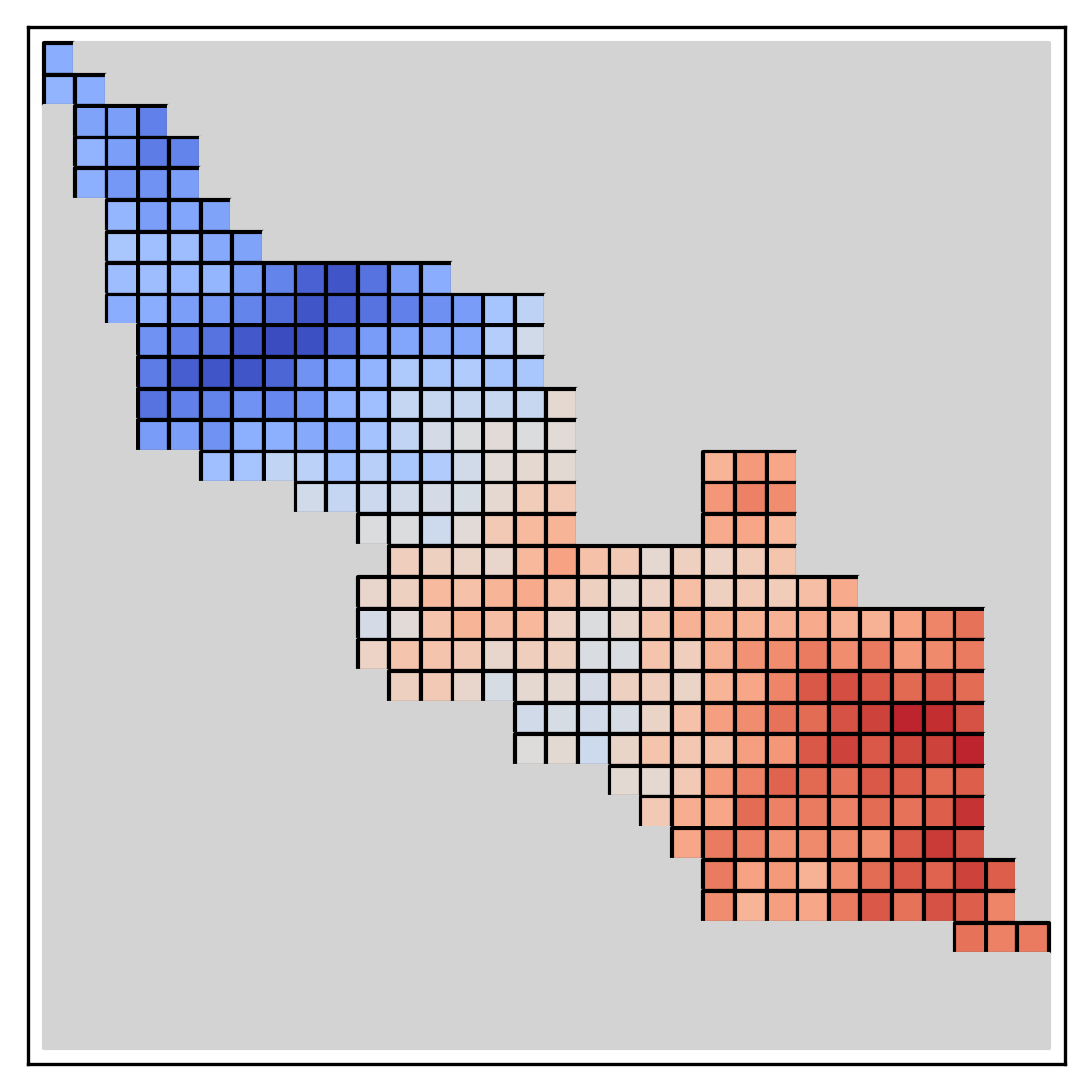}
    \label{Fig:1d}
    }
    \caption{\small (a) Full Environment {\small ($\M$)}. (b) Partially Explored Map {\small ($\M_e$)}. (c) A representative Data Transfer Map {\small ($X_\mathcal{T}$)}. (d) Corresponding Information Map {(\small $X_{\mathcal{I}}$}).}
    \label{Figure: 2} 
    \vspace{-4mm}
\end{figure}
A given agent $\id$ has a limited sensing range of $n^\id \times n^\id$ cells centered on its current position, where $n^\id<N$.
We have two types of agents: helper (i.e., UAV) and receiver (i.e., UGVs).
We use $\id = \h$ to denote the quantities corresponding to the helper and $\id = \s$ for a generic receiver agent.  
As agents traverse and explore the environment, the occupancy values of the corresponding cells are updated.
At any time~$\tm$, the explored and unexplored portion of the map for agent $\id$ are denoted by $\M^\id_e(\tm)$ and $\M^\id_u(\tm)$, respectively.
Figure~\ref{Fig:1b} illustrates the known and unknown regions on a partially explored map.
The gray, edgeless regions represent unexplored areas, while the sections containing cells with varying occupancy values correspond to the explored regions.

The helper agent transmits map observations from its explored region $\M_e^{\h}(\tm)$ to a receiver agent in the form of cell occupancy values and their coordinates.
Let $\X^{\s}_{\calg{T}}(\tm) \subseteq \M^{\h}_e(\tm)$ denote the set of cell observations transmitted up to time $\tm$ to receiver $\s$.  
Figure~\ref{Fig:1c} illustrates an example map showing the transmitted cells.
Grid cells that are explored but not transmitted are defined as \textit{informative} cells, and they are represented by the set $\X^{\s}_{\calg{I}}(t) = \M^{\h}_e(\tm) \setminus \X^{\s}_{\calg{T}}(\tm)$.  
The information map in Figure~\ref{Fig:1d} highlights the locations of these non-transmitted informative cells, where different color shades represent varying levels of information intensity (details in Section~\ref{subsec:VoI}).
Note that the transmitted data set and the informative sets can differ for each receiving agent, since the helper sends customized map portions based on the task relevancy (e.g., predicted path) of those agents.

\subsection{Problem Scope and Assumptions}

In this framework, we consider a heterogeneous team consisting of a single helper and a set $\boldsymbol{R}\!=\!\{1,2,\ldots,r\}$ of $r$ receiver agents operating in a deterministic but unknown environment.
The helper and receiver agents are formally hereafter referred to as the `Supporter' and `Seeker', respectively.
Each agent is equipped with a limited range ($n^\s \times n^\s$; $\s \in \boldsymbol{R}$) local sensing device  to observe its surroundings and reach its goal with minimum path cost.  
The supporter agent, on the other hand, is modeled as an aerial drone performing reconnaissance tasks at a high altitude without worrying about the ground obstacles.  
The supporter UAV is assumed to have a bigger sensing region ($n^\h \times n^\h$), where $n^\h > n^\s$. 
The supporter assists the seekers by transmitting relevant map observations from its explored region to help them navigate.  
By default, the supporter explores along a predefined path; however, upon receiving a request from a seeker, it adapts its behavior to explore judiciously based on the seekers' way-points.

It is assumed that at least one feasible path exists in the environment for a seeker to reach its goal, ensuring that, with exhaustive local sensing and potentially a longer path, the seeker can still navigate to its goal even without support.
All agents share the grid values \(\phi^{\text{obs}}\) and \(\phi^{u}\) as common knowledge.
The available bandwidth \(B\) for data transmission is assumed to be constant throughout the environment. 

\subsection{Problem Setup} \label{subsec:problemSetup}

At time \(\tm\), each seeker agent in the set of active seekers, denoted by \(\boldsymbol{R_a}(\tm) \subseteq \boldsymbol{R}\), navigates from its start location \(\mathbf{S}^\s\) to its goal location \(\mathbf{G}^\s\) using the action set \(U^{\s} = \{ \uparrow, \downarrow, \leftarrow, \rightarrow \}\).
Active seekers are those agents that have not yet reached their goal positions, formally defined as:
$\boldsymbol{R_a}(\tm) = \{ \s \in \boldsymbol{R} \mid \mathbf{p}^\s(\tm) \neq \mathbf{G}^\s \}$.
The supporter agent initially explores along its default path, e.g., a boustrophedon path.
Once the supporter receives a request from any seeker agent in the form of way-points (details in \eqref{eq:path_data}), it proceeds to explore those received way-points from its current position \(\mathbf{p}^\h(\tm)\).
The supporter has a larger action set \(U^{\h} = \{ \uparrow, \downarrow, \leftarrow, \rightarrow, \searrow, \nearrow, \swarrow, \nwarrow \}\) since it is an aerial agent and all eight neighboring cells are traversable.
The formulation readily extends to the case when the supporter's action set is limited or the some of the cells are untraversable.\\

\noindent\textbf{Problem Statement: } The supporter agent receives requests from multiple active seeker agents. 
A key challenge for the supporter is to determine its exploration strategy in order to gather relevant information efficiently. 
Additionally, communication at every time-step is limited by a bandwidth constraint, which requires the supporter to decide \textit{which portions} of the available information, and \textit{how much} of it, should be transmitted to each seeker agent to enhance their navigation performance.
We also investigate the \textit{communication-navigation trade-off}: the amount of data communicated by the supporter versus the total navigation cost incurred by the seekers.


\section{Proposed Framework} \label{Proposed Framework}

\begin{figure}
\centering
\includegraphics[width = \linewidth]{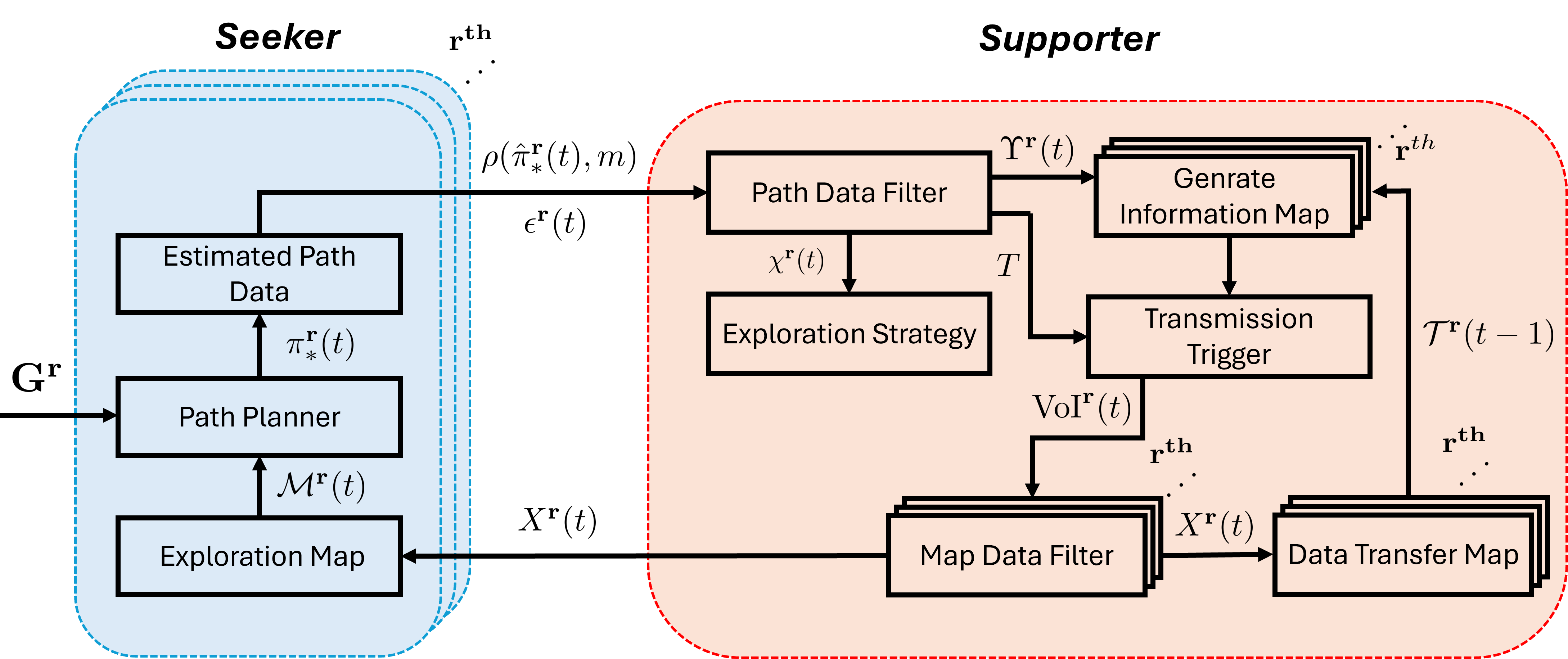}
\caption{\small Flowchart illustrating the operation at time \(t\). Active seekers send their path information to the supporter. Based on the received path data ({\small $\rho(\hat\pi^\s_*(\tm),m)$}), the supporter updates their corresponding \textit{information map} to select relevant data for transmission and determines the \textit{exploration path} to gather additional information.}
\label{Figure: 3}
\vspace{-4mm}
\end{figure}

The proposed framework is outlined in Figure~\ref{Figure: 3}.  
The supporter addresses each seeker’s request by dividing it into two parts—one that can be fulfilled using the available information and another that requires additional exploration.  
For the portion where information is already available, a \textit{value-of-information}-based approach is used to select the most relevant data to transmit, followed by a mixed-integer programming-based bandwidth allocation scheme.
To gather new observations, an agent-aware informative exploration strategy is employed.
This section describes each module of the proposed framework in detail.

\subsection{Path Planner}

Both the seeker and supporter agents have distinct traversal costs based on their respective models and action sets, as described in~\Cref{subsec:problemSetup}.  
For the seeker agent, the cost of moving to the next cell is proportional to its occupancy value, and cells with occupancy values greater than \(\phi^{\text{obs}}\) are considered non-traversable.  
For path planning, the seeker assumes the occupancy value of unexplored cells to be \(\phi^{u}\).  
The cost of traversing from a  cell \(\p\) to a neighboring cell \(\p'\) for the seeker agent is defined as  
\begin{equation} \label{eq: seeker cost of traversing neighbour vertex}
c^{\s}(\p,\p') =
\begin{cases} 
o_{\p'} + \lambda_{1}, & \text{if } o_{\p'} \leq \phi^{\text{obs}}, \\
\infty, & \text{otherwise},
\end{cases}
\end{equation}  
where \(\lambda_{1}\) is a constant representing the lateral edge cost of moving from \(\p\) to \(\p'\).  
The cost formulation in~\eqref{eq: seeker cost of traversing neighbour vertex} ensures that the seeker avoids paths passing through untraversable~cells.

In contrast, the supporter’s movement is not affected by cell occupancy.  
Its traversal cost depends only on the edge cost between \(\p\) and \(\p'\), defined as :
\begin{equation} \label{eq: supporter cost of traversing neighbour vertex}
c^{\h}(\p,\p') = \lambda_{2}.
\end{equation} 
Since the supporter’s action set \(U^{\h}\) allows both lateral and diagonal movements, we allow  
\(\lambda_{2}\) to take two possible values: \(\lambda_{2} = \lambda_{1}\) for a lateral move, and \(\lambda_{2} = \lambda_{1}\sqrt{2}\) for a diagonal move.

For a generic agent \(\id\), the cost of traversing its path \(\pi^\id \triangleq \{ \p^{\id}(1), \p^{\id}(2), \ldots, \p^{\id}(\ell) \}\) is given by:
{\small
\begin{align} \label{eq: total path cost}
    \mathcal{C}^\id[\pi^\id] = \sum_{\tm=1}^{\ell-1} c^\id(\p^{\id}(\tm), \p^{\id}(\tm+1)),
\end{align}}where, depending on the agent type \(\id\), the cost function \(c^\id\) corresponds to either \eqref{eq: seeker cost of traversing neighbour vertex} or \eqref{eq: supporter cost of traversing neighbour vertex}.

Let the set of all feasible paths for agent \(\id\) from an arbitrary start position \(\mathbf{S}^\id\) to a goal position \(\mathbf{G}^\id\) be denoted by \(\Pi^\id(\mathbf{S}^\id,\mathbf{G}^\id)\). 
Then, the optimal path for agent \(\id\) is defined~as: 
\begin{equation} \label{eq: optimal path}
{\pi}^{\id}_*(\mathbf{S}^{\id},\mathbf{G}^{\id}) = 
\argmin_{\pi \in \Pi^{\id}(\mathbf{S}^{\id},\mathbf{G}^{\id})} \mathcal{C}^\id[\pi].
\end{equation}  
When the map of agent \(\id\) is not fully explored, its predicted optimal path at time \(\tm\) is denoted by \(\hat{\pi}^\id_*(\mathbf{S}^{\id},\mathbf{G}^{\id},\tm)\).
With a slight abuse of notation, we use \(\hat{\pi}^\id_*(\tm)\) to represent the seeker’s estimated path whenever the start and goal positions are clear from the context. 

\subsection{Communication of Map and Path Data}

At time $\tm$, a seeker’s estimated optimal path is denoted by $\hat\pi_*^{\s}(\tm) \triangleq \{ \p^{\s}(\tm), \p^{\s}(\tm+1), \ldots, \p^{\s}(\tm+\ell(\tm))\}$, where $\p^{\s}(s) \in \M$ represents the seeker’s location at time $s \ge \tm$.  
The seeker's predicted path is of length $\ell(\tm)$, with the endpoint $\p^{\s}(\tm+\ell(t))$ corresponding to the desired goal position $\textbf{G}^{\s} \in \M$.
Note that some of the points $\p^{\s}(k)$ may go through unexplored cells and therefore the path may not be feasible. 
This motivates a seeker to request occupancy information from the supporter for enhancing its situational awareness and computing a better~path.

At each time $\tm$, the seeker agent sends its estimated path to the supporter. 
Path way-points that fall within the seeker's already explored/known portion of the map (i.e.,~within {\small\(\M_e^{\s}(\tm)\)}) are excluded from this set. 
Also, instead of sending all the unknown way-points, it sends a subset~of~them:
{\small \begin{align} \label{eq:path_data}
    \rho({\hat\pi}^{\s}_*(\tm),m) \triangleq \{ \p^{\s}(\tm+mk) \}_{k = 1}^{\lfloor \frac{\ell(\tm)}{m} \rfloor} \setminus \M^{\s}_e(t).
\end{align} }The parameter \(m\) determines the path sampling interval; when \(m = 1\), the seeker sends all unexplored path locations as way-point candidates to the supporter.

In addition to the way-points set, the seeker also transmits a \textit{path-uncertainty} parameter representing the  fraction of its estimated path that lies within the unexplored region.  
The \textit{path-uncertainty} parameter at $\tm$ is illustrated in the bottom-right of Figure~\ref{Figure: 1} and mathematically described as:
\begin{equation} \label{eq:seeker estimation cell ratio}
    \epsilon^{\s}(\tm) = \frac{| \hat\pi_*^{\s}(\tm) \cap \M^{\s}_u(\tm)|}{| \hat\pi_*^{\s}(\tm)|} \in [0,1],
\end{equation}
where $|\cdot|$ denote the cardinality of a set.
Therefore, if the entire path of the seeker $\s$ is within the unexplored region $\M^{\s}_e$, then  $\epsilon^\s(\tm) = 1$---representing the fact that the path is \textit{completely uncertain} and occupancy information from the supporter can highly help in updating the current path. 
On the other hand, $\epsilon^\s(t) = 0$ indicates that the current path of the seeker passes through its known region and information from the supporter is unlikely to improve the current path.

In return, the supporter transmits path-relevant occupancy information from its explored map to the requesting seekers.  
The supporter transmits this information at a fixed periodic interval $T$, and the amount of data transmitted is constrained by the available bandwidth $B$. 
For map data transmission, the supporter encapsulates grid information in a tuple format containing the occupancy value and corresponding grid coordinates, denoted as $(o_{\p}, x, y)$, where $\p = (x, y) \in \M^{\h}_e(\tm)$.  
Let $b_0$ denote the required number of bits for the transmission of a single cell’s information $(o_{\p}, x, y)$.
Consequently, a total of at most $\tfrac{B}{b_0}$ cell information can be transmitted at any given communication instance.

\subsection{Supporter's Exploration Strategy} \label{subsec:exploration}

Supporter exploration determines which trajectory to follow based on the selected way-points for subsequent exploration.  
As described in~\Cref{subsec:problemSetup}, the supporter agent follows one of two paths: a predefined path (e.g., lawn-mower path) or a utility-based, agent-aware, strategic exploration path soon to be defined.  
Unless the supporter agent receives a request from the seeker group, it periodically explores along its default set of way-points \(\{\p^\h_1, \p^\h_2, \p^\h_3, \dots, \p^\h_d\}\), where $\p^\h_d = \p^\h_1$ to ensure a periodic path. 
The resultant trajectory followed by the supporter is referred to as its \textit{Default Path}, denoted by~\(\pi^\h_{\rm default}\).

The supporter must explore the environment based on the path data \(\rho(\hat{\pi}^\s_*(\tm), m)\), for $\s \in \boldsymbol{R_a}(\tm)$, received from the active seeker agents at time $t$. 
For a strategic exploration, the supporter first identifies the way-points within its explored region \(\M^\h_e(\tm)\) for which information is already available to transmit.
The remaining way-points that lie within the unexplored region, \(\M^\h_u(\tm)\), are considered potential candidates for gathering new information.
Let \(\chi^\s(\tm)\) denote the set of filtered way-points for the \(\s^{\text{th}}\) active seeker at time \(\tm\): 
\begin{align} \label{eq:set of unexplord way points}
    \chi^\s(\tm) =\rho(\hat{\pi}^\s_*(\tm), m) \cap \M^\h_u(\tm).
\end{align}

The received way-points from each seeker are ordered from the seeker's current position to its goal $\mathbf{G}^\s$.
Let a generic set of $v$ received way-points be denoted as $\{\mathbf{q}_1, \dots, \mathbf{q}_v\}$, with $\mathbf{q}_v$ being the way-point nearest to goal $\mathbf{G}^\s$.
We define an agent-specific, path-aware, utility function:
{\small\begin{align} \label{eq:utility score of waypoints}
    \mathcal{U}^\s(\tm) = w_1 {\epsilon}^\s(\tm) + w_2 \frac{\lVert \mathbf{q}_v - \mathbf{q}_1 \rVert_2}{\lVert \mathbf{q}_v - \mathbf{p}^\h(\tm) \rVert_2},
\end{align}}where $w_1,w_2\ge 0$ are hyperparameters, \({\epsilon}^\s(\tm)\) is the estimated \textit{path-uncertainty fraction} defined in \eqref{eq:seeker estimation cell ratio}, and the terms
\(\lVert \mathbf{q}_v - \mathbf{q}_1 \rVert_2\) and
\(\lVert \mathbf{q}_v - \mathbf{p}^\h(\tm) \rVert_2\) represent the Euclidean distances between the first and last way-points, and between the last way-point and the supporter’s current position $\p^\h(\tm)$.

The first term of the utility function encapsulates the seeker's need through the path-uncertainty fraction whereas the second term aims to encapsulate whether the supporter will be able to reach and gather the path-specific data before the seeker itself reaches there.  
The supporter will explore the way-points with the highest utility score:
\begin{align} \label{eq:set with max utility}
    \chi^{*}(\tm) = {\arg\max}_{\s \in \boldsymbol{R_a}(\tm)}~ \mathcal{U}^{\s}(\tm).
\end{align}

The supporter visits the filtered set of unexplored way-points in the reverse order. 
Visiting the way-points in reverse allows the supporter to gather remote observations farther away from the seeker’s local sensing region.
Consequently, the supporter's path to explore from its current location is given by:
\begin{align} \label{eq:supporter concatenated path}
    \pi_{\chi^*}^{\h}(\tm) \!=\! \pi_{*}^{\h}(\p^\h(\tm), \q_v) \circ \pi_{*}^{\h}(\q_v, \q_{v-1}) \circ \cdots \circ \pi_{*}^{\h}(\q_2, \q_1),
\end{align}
where \(\circ\) denotes the path concatenation operation, and recall that \(\pi_{*}^{\h}(\p, \p')\) represents the supporter’s optimal path from cell $\p$ to $\p'$, as defined in \eqref{eq: optimal path}.

At each time \(\tm\), the supporter agent thus determines how to explore based on the path data \(\rho(\hat{\pi}^\s_*(\tm), m)\) received from the active seeker agents. 
The supporter either continues along the same path from the previous timestep \(\pi^\h_{\chi^*}(\tm-1)\), explores along a newly generated path \(\pi^\h_{\chi^*}(\tm)\), or follows its \textit{default path} defined by the way-points \(\pi^\h_{\rm default}\).  
The supporter’s exploration path at time \(\tm\) is given by:
\begin{equation} \label{eq:supporter exploration}
\pi^\h(t) =  
\begin{cases} 
\pi^\h_{\rm default}, & \text{if } \chi^*(\tm) = \emptyset, \\
\pi^\h_{\chi^*}(\tm-1), & \text{if } \chi^*(\tm) = \chi^*(\tm-1), \\
\pi^\h_{\chi^*}(\tm), & \text{if } \chi^*(\tm) \ne \chi^*(\tm-1). \\
\end{cases}
\end{equation}
Note that if the supporter receives no path data from any seeker  \((\rho(\hat{\pi}^\s_*(\tm), m) = \emptyset)\), or if all received way-points from the seekers lie within the supporter’s explored map, it defaults to following its predefined path.
That is, once the supporter completes a exploration path $\pi^\h_{\chi^*}$, it goes to the closest way-point on its default path and continues the default motion.

\subsection{Supporter's Information Map: Value-of-Information} \label{subsec:VoI}

Based on the path data received from multiple seeker agents, the supporter determines $\textit{what to send}$ to each seeker according to their respective needs.
The supporter uses the \textit{information map} to determine which piece of information to transmit in response to each seeker request.
A separate information map is maintained for each seeker agent to capture non-redundant observations that reflect the variance between the assumed prior and the acquired map data.
The supporter uses the received path data, $\rho(\hat\pi^\s_*(\tm),m)$, to update the information map corresponding to the $\s^{th}$ seeker, incorporating its data-transfer map and region-of-interest (RoI) weights, where each cell's information magnitude denotes its likelihood of transmission.

\subsubsection{Data Transfer Map}

For each seeker agent, the supporter maintains a data-transfer map that tracks the map-data transmitted to that agent.
The occupancy value for cell location $\p$ in the data transfer map of $\s^{th}$ seeker at time $\tm$ is denoted by $\mathcal{T}^\s_{\p}(\tm)$, and given by:
{\small\begin{equation} \label{eq: Data transfer value}
\mathcal{T}^\s_{\p}(\tm) =  
\begin{cases} 
o_{\p}, & \text{if \({\p} \in  \X_{\mathcal{T}}^\s(\tm)\), }\\
 \phi^{u}, & \text{otherwise},
\end{cases}
\end{equation}}here, recall that $\X_{\mathcal{T}}^\s(\tm)$ is set of data transfer map at time~$\tm$ defined in~\Cref{subsec:prelim}.
The occupancy values for the transmitted cell locations are set to the true occupancy $o_\p$ in $\mathcal{T}$, whereas the unsent ones are set to $\phi^u$, reflecting the seeker's prior that unobserved cells have an occupancy value~of~$\phi^u$.

The difference $(\mathcal{T}^\s_{\p}(\tm) - o_\p)$ represents the deviation between the supporter's observed and the seeker's prior occupancy.
A large difference indicates a poor occupancy estimate, unless the seeker has explored it and the supporter does not know.
Figure~\ref{Fig:4b} shows the difference \((\mathcal{T}^\s_{\p}(\tm) - o_\p)\), whereas Figure~\ref{Fig:4a} shows \(\M^\h_e(\tm)\) along with the received way-points.
Note that red and blue denote negative and positive differences, respectively; white indicates zero.

\subsubsection{Region of Interest (RoI)}

The region of interest emphasizes on areas of the information map that are most relevant to the given  seeker.
Cells with large difference values \((\mathcal{T}^\s_{\p}(\tm) - o_\p)\) do not always imply navigational relevance.
While cells with smaller differences located near the seeker’s path may, in fact, carry more useful information. 
Therefore, each cell’s difference (information) value is scaled by its proximity to the seeker’s estimated path using an \textit{RoI-filter}.

Supporter extracts the set of way-points that lie within the explored portion of its map for $\s^{th}$ seeker, given by $\Upsilon^\s(\tm) =\rho(\hat{\pi}^\s_*(\tm), m) \cap \M^\h_e(\tm)$.\footnote{Note that \(\Upsilon^\s(\tm) = \rho(\hat\pi^\s_*(\tm),m) \setminus \chi^\s(\tm)\).} 
The interest (RoI) value for given cell location $\p$ on map $\M$ for $\s^{th}$ seeker at time is denoted by $\Dot{\iota}^\s_\p(\tm)$ and defined as:
\begin{equation} \label{eq:interest value}
\Dot{\iota}^\s_\p(\tm) = \beta +
\sum_{\q \in \Upsilon^\s(\tm)} \alpha_\q e^{-\frac{\|\p - \q\|^{2}}{2\sigma^{2}}} .
\end{equation}
Here, \(\alpha_{\q}\) is chosen to be proportional to the sequential order of the way-point \(\q\); that is, \(\alpha_{\q} = \alpha k\) for the \(k^{\text{th}}\) way-point in \(\Upsilon^{\s}\).
The parameter \(\beta > 0\) represents the baseline interest and \(\sigma\) defines the decay-rate (width) of the region of interest.
In this approach, more weight is assigned to the first way-point (i.e., the one closest to the seeker's location) as that is the most needed information in the current time.

\begin{figure}
    \centering
    \subfigure[]{
    \includegraphics[width=0.23\linewidth]{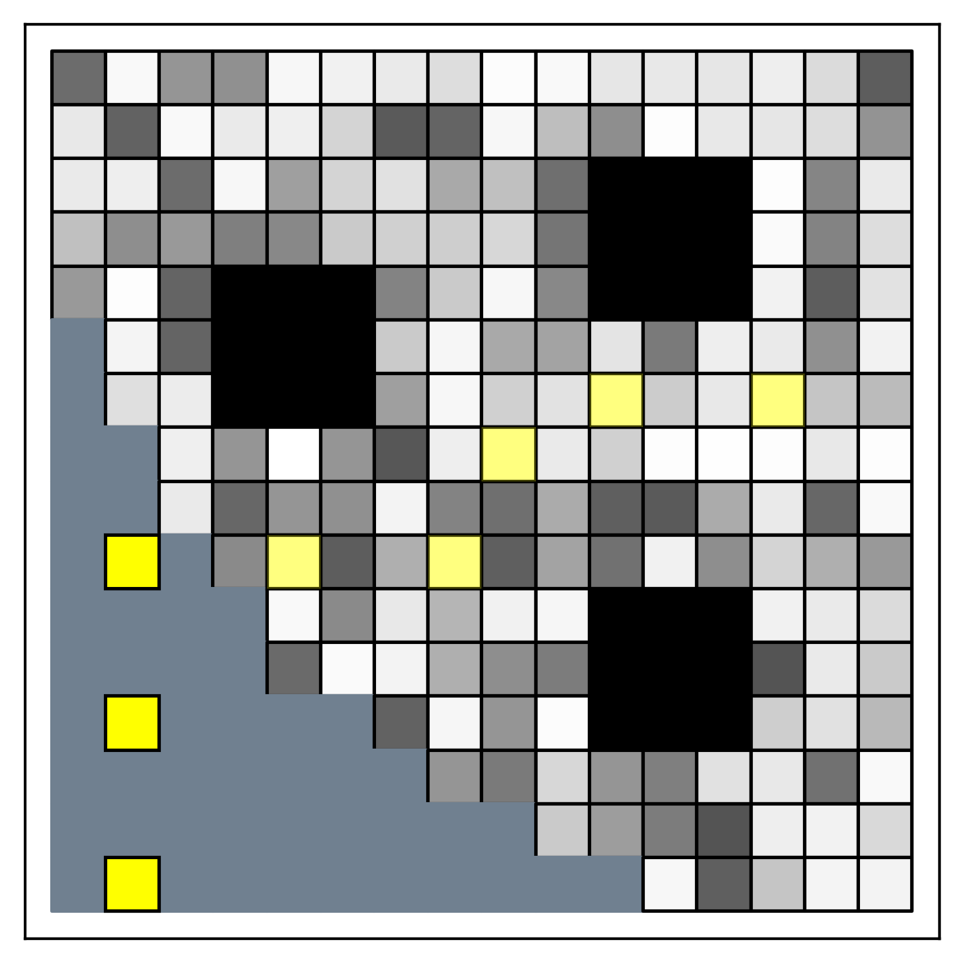}
    \label{Fig:4a}
    } \hspace{ -5mm}
    \subfigure[]{
    \includegraphics[width=0.23\linewidth]{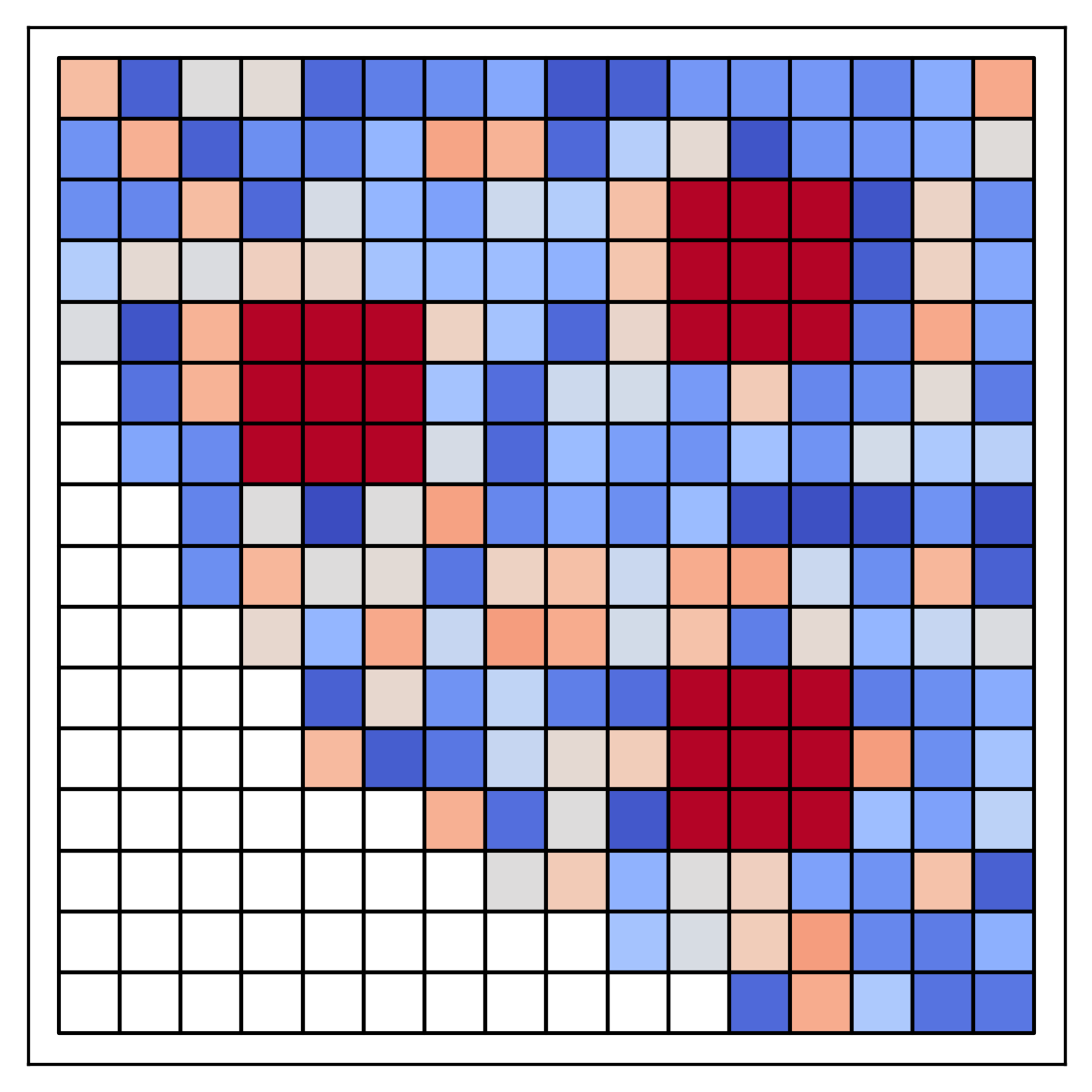}
    \label{Fig:4b}
    } \hspace{ -5mm}
    \subfigure[]{
    \includegraphics[width=0.23\linewidth]{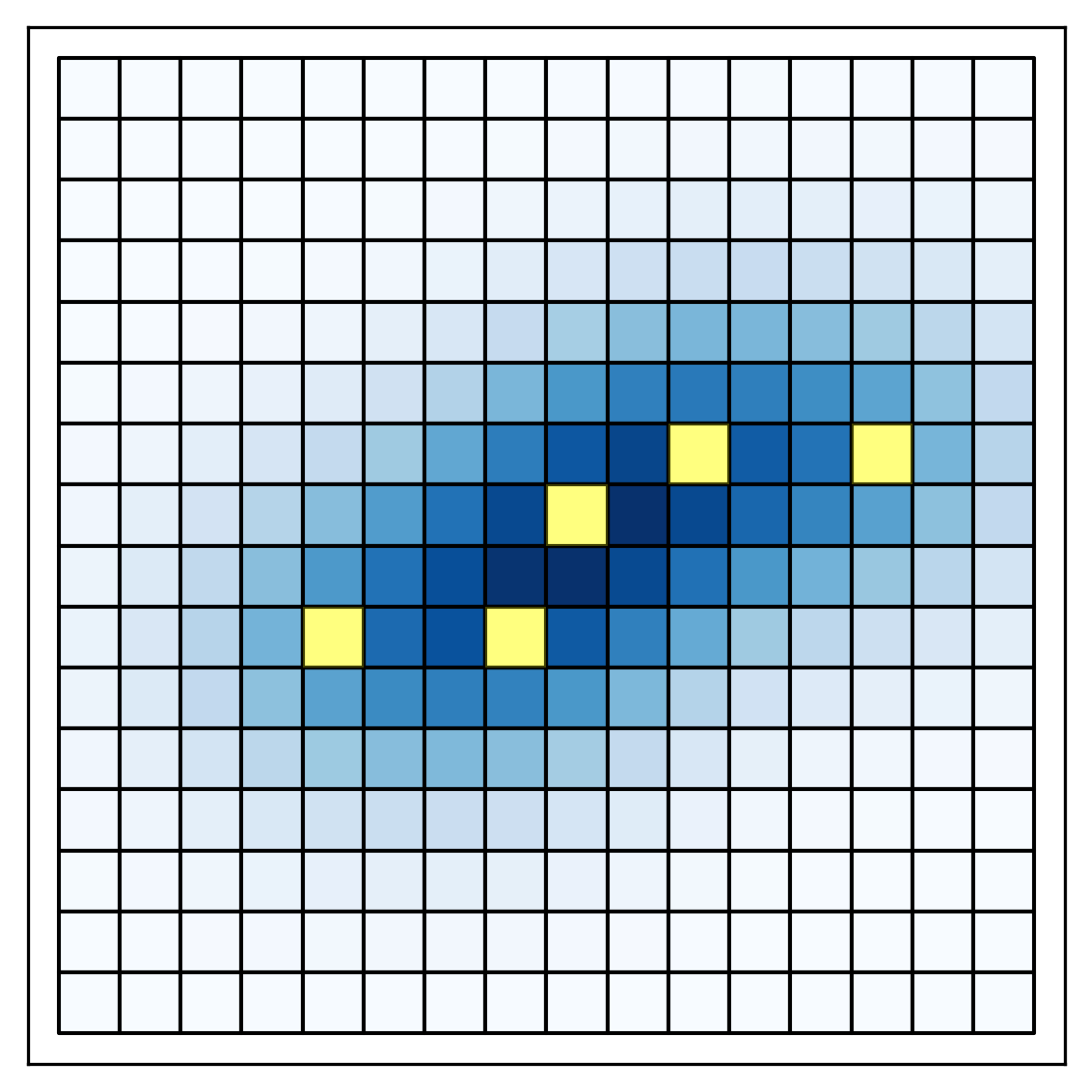}
    \label{Fig:4c}
    } \hspace{ -5mm}
    \subfigure[]{
    \includegraphics[width=0.23\linewidth]{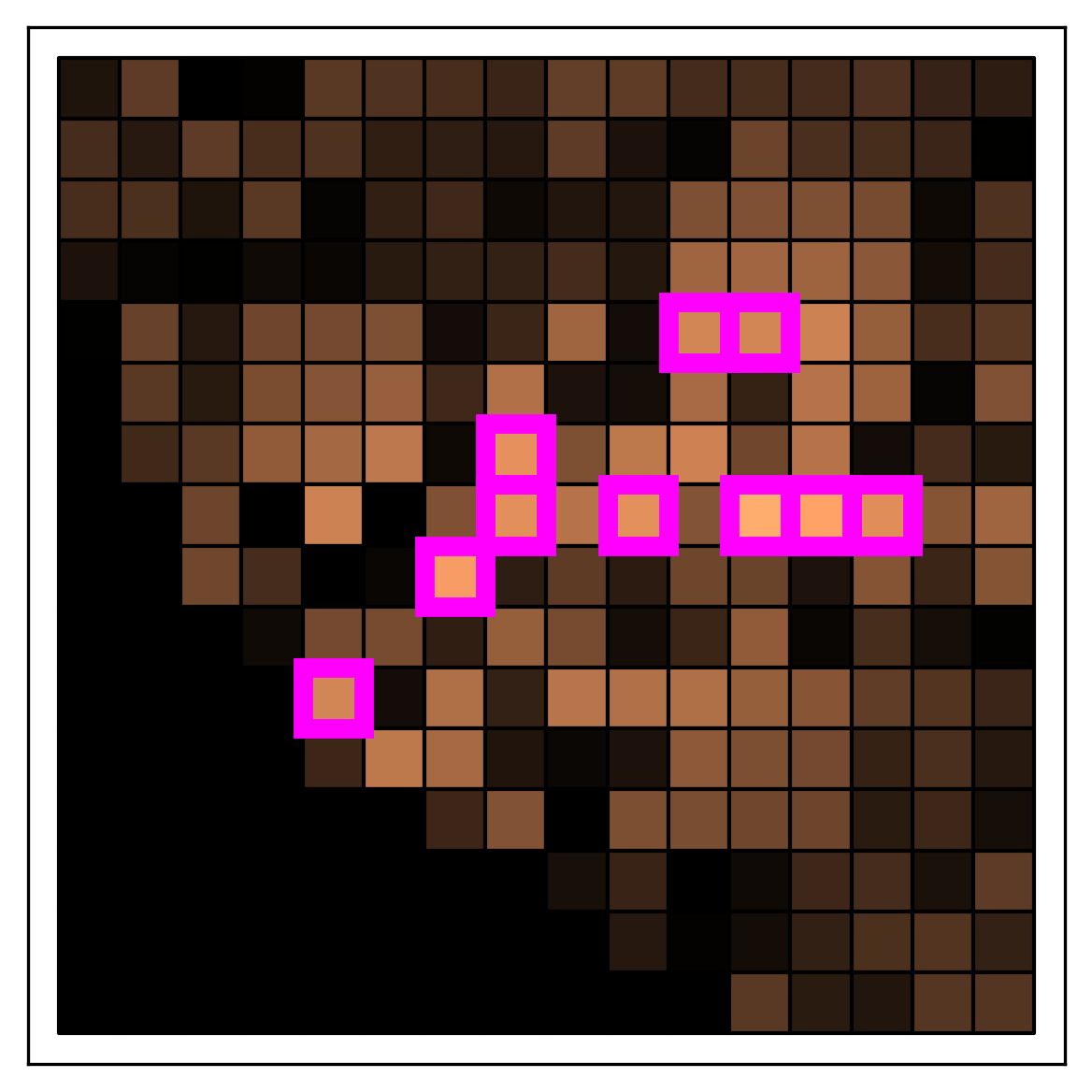}
    \label{Fig:4d}
    }
    \caption{\small (a) Supporter Exploration Map \(\mathcal{M}^\h(\tm)\): the slate-gray region represents the unexplored area \(\mathcal{M}^\h_u(t)\). The yellow cells correspond to the $\s^{th}$ seeker’s path data \(\rho(\hat{\pi}^\s_*(\tm), m)\), sampled at \(m = 3\). (b–d) Maps corresponding to the \(\s^{\text{th}}\) seeker agent:  (b) Supporter Difference Map \((\mathcal{T}^\s_{\p}(\tm) - o_\p)\): red indicates a positive difference, blue a negative one, and white no difference.  (c) Region of Interest Map \(\Dot{\iota}^\s_\p(\tm)\): darker shades represent higher interest values. (d) Weighted Information Map \(\mathrm{VoI}^\s_{\p}(\tm)\): cells with magenta boundaries denote the top \(10\) cells with most \textit{value-of-information}.} 
    \label{Figure 4} 
    \vspace{-4mm}
\end{figure}

Using the \textit{region-of-interest} and \textit{data-transfer} map for the $\s^{th}$ seeker,  the supporter computes the corresponding information map as follows:
\begin{equation} \label{eq:value-of-information}
\mathrm{VoI}^\s_{\p}(\tm) = 
\Dot{\iota}^\s_\p(\tm) \left( \mathcal{T}^\s_{\p}(\tm-1) - o_\p\right),
\end{equation}
where \((\mathcal{T}^\s_{\p}(\tm-1) - o_\p)\) denotes the information difference after time $\tm-1$ and before transmission at time $\tm$. 
For each cell $\p$ and seeker $\s$, $\mathrm{VoI}^\s_{\p}(\tm)$ represents the \textit{value-of-information} (VoI) of that particular cell to that specific seeker at the given time $\tm$, capturing both the time-dependent quality \((\mathcal{T}^\s_{\p}(\tm-1) - o_\p)\) and the relevance, \(\Dot{\iota}^\s_\p(\tm)\), of that information to the seeker’s task.
Figure~\ref{Fig:4c} illustrates the resulting region of interest, derived from the received light-yellow way-points within the explored area of the supporter’s map.

\subsection{Supporter's Agent- and Task-Aware  Bandwidth Allocation} \label{Channel Bandwidth Allocation}

In a team where a supporter agent assists multiple seekers, it must decide how to allocate an appropriate portion of the bandwidth to each seeker agent in a dynamic and adaptive manner. 
To achieve this, it must utilize the available bandwidth efficiently by regulating the amount of information transmitted to each seeker at a given time \(\tm\).
%

\begin{figure}
    \centering
    \includegraphics[width=0.5\linewidth]{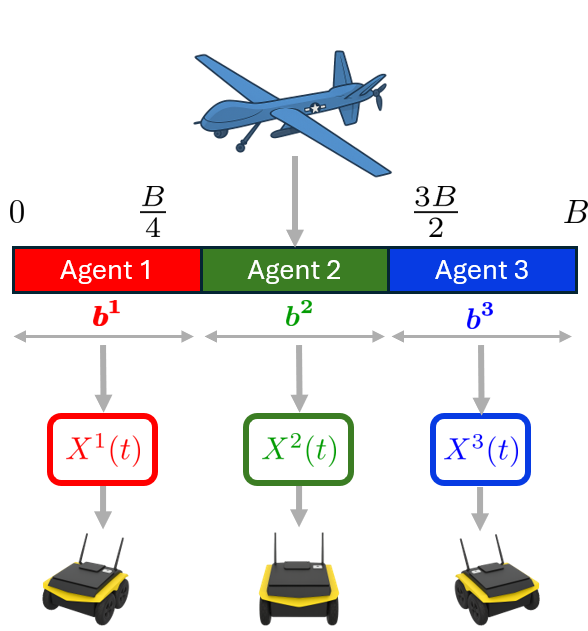}
    \caption{\small Bandwidth allocation for a single-supporter and three-seeker team. The allocated bandwidths for each agent are shown by the red, green, and blue sections, representing \(\boldsymbol{b}^1\), \(\boldsymbol{b}^2\), and \(\boldsymbol{b}^3\), respectively.}
    \label{Figure: 5}
    \vspace{-4mm}
\end{figure}

Figure~\ref{Figure: 5} illustrates a supporter agent assisting three seekers by allocating a portion of the available bandwidth to each seeker.  
Here the supporter agent allocates its bandwidth resources to each active seeker based on the amount of \textit{informative cells} it has available to transmit. 
The set of informative cells at time $\tm$ for the $\s^{th}$ active seeker agent is defined as $\mathcal{D}^\s(t) = \{\p ~|~ |\mathrm{VoI}^\s_{\p}(\tm)| > 0\}$, where $|\mathcal{D}^\s(\tm)|$ denotes the total number of such informative cells available for transmission.
To formulate an generic bandwidth allocation problem for all seeker agents at time $\tm$, let \(\mathbf{d} = [|\mathcal{D}^1(\tm)|,|\mathcal{D}^2(\tm)|,|\mathcal{D}^3(\tm)|,\ldots,|\mathcal{D}^r(\tm)|]\) and \(\mathbf{e} = [\epsilon^1(\tm), \epsilon^2(\tm), \epsilon^3(\tm), \ldots, \epsilon^r(\tm)]\) be the vectors representing the amount of informative cells possessed by the supporter agent and the path-uncertainty estimates for the corresponding seeker, respectively.  
Here, $r = |\boldsymbol{R_a}(\tm)|$ denotes the number of active seekers.
To transmit data to each seeker at a given time instant, let \(\mathbf{b} = [\boldsymbol{b}^1(\tm), \boldsymbol{b}^2(\tm), \boldsymbol{b}^3(\tm), \ldots, \boldsymbol{b}^r(\tm)]\) denote the vector representing the amount of bandwidth to be allocated to the $r$ seekers.
To determine the optimal bandwidth allocation vector $\mathbf{b}$, the following mixed integer-linear programming (MILP) method is employed:

{\small\begin{equation} \label{eq: Bandwidth channel allocation}
\begin{split}
\begin{aligned}
& \max_{\mathbf{b}}   && \mathbf{e}^\intercal \mathbf{b} \\
& \text{s.t.}   && \sum_{\s=1}^{r} \boldsymbol{b}^\s(t) \leq B, \\
&                     && \boldsymbol{b}_{\min} \le \boldsymbol{b}^\s(t) \le|\mathcal{D}^\s(t)|, \quad  \boldsymbol{b}^\s(t) \in \mathbb{Z}^{+},\quad \forall\s\in \boldsymbol{R_a}.\!\! \\
\end{aligned}
\end{split}
\end{equation}}Based on the above formulation, the amount of channel allocation for a given seeker depends on two parameters: the number of available informative cells $|\mathcal{D}^\s(\tm)|$ and the \textit{path-uncertainty fraction} $\epsilon^\s(\tm)$.  
The path-uncertainty value $\epsilon^\s(\tm)$ scales the decision variable $\boldsymbol{b}^\s(\tm)$ proportionally to the level of assistance required by the \(\s^{th}\) seeker (i.e., a higher value of $\epsilon^\s(\tm)$ indicates that the $\s^{th}$ seeker requires more urgent support).  
The first constraint in~\eqref{eq: Bandwidth channel allocation} states that the cumulative sum of allocated bandwidth across all seekers must not exceed the total channel bandwidth limit~\(B\).  
The second constraint guarantees that, for \( \mathbf{r}^{\text{th}} \) seeker, the allocated bandwidth \( \boldsymbol{b}^\s(\tm) \) remains within the range defined by the minimum bandwidth allocation constant $\boldsymbol{b}_{\min}$ and the number of informative cells available to the supporter for that seeker.
The objective function, which is a weighted sum of the decision variables, is maximized to determine the optimal bandwidth allocation for each seeker.

The set $\mathcal{D}^\s(\tm)$ for seeker \(\mathbf{r}\) is computed independently of the allocated bandwidth $\boldsymbol{b}^\s$ for that agent and may, at times, contain more cells than can be transmitted at time~\(\tm\).  
To satisfy the bandwidth constraint, the cells are sorted by priority (VoI), and the first \(\tau = \boldsymbol{b}^\s / b^0\) cells are selected for transmission to seeker \(\mathbf{r}\), forming the set \(X^\s(\tm)\).  
The cumulative transmitted set is then updated as \(X^\s_{\mathcal{T}}(\tm) = X^\s_{\mathcal{T}}(\tm-1) \cup X^\s(\tm)\), where \(\boldsymbol{b}^\s\) denotes the bandwidth allocated to seeker \(\s\).  
Figure~\ref{Fig:4d} highlights the selected transmission cells \(X^\s(\tm)\) in bold magenta.

\section{Simulation Experiments} \label{Simulation Results}

This section describes the simulation environment, the occupancy grid map(s), and the selection of hyperparameter values used for experimentation.
It also discusses the baseline methods and evaluation metrics employed to compare the performance of the proposed framework. 
\begin{mdframed}[backgroundcolor=gray!10, roundcorner=10pt,
  innerleftmargin=4pt, innerrightmargin=4pt,
  innertopmargin=4pt, innerbottommargin=4pt]
  \small
\textbf{Simulation Parameters}:\\
\(\phi^{\min} = 0\), \(\phi^{\max} = 100\),  \(\phi^{\text{obs}} = \phi^u = 50\), $m=3$, $n^\s=3$\\ \hspace{25mm} 
$w_1=0.4$, $w_2=0.6$,
$\boldsymbol{b}_{\min} = \frac{{\epsilon}^\s(t) \, B}{|\boldsymbol{R}|}$, $\beta = 1, T=1, \alpha=1000$.\\ [2pt]
\end{mdframed}

\paragraph{Baseline Methods} 

To evaluate the performance of the proposed method, we compare it with two baseline approaches.
The first is the \textit{Fully Informed (FI)} method, where the supporter communicates \textbf{all} of its local map observations at every time step, and each seeker receives all observed information instantaneously.
This is done without any bandwidth constraint.
This method ensures that each seeker knows all the information that the supporter has gathered.
This method is expected to give the optimal seeker navigation cost at the expense of the highest communication~overhead.

The second baseline is the \textit{Uninformed (UI)} method, where there is no communication from the supporter to the seekers, and each seeker navigates to its goal independently.
This method demonstrates the worst-case seeker navigation due to the lack of communication.
These two baseline methods highlight the unique communication–navigation balance that our method is able to achieve.

Additionally, for the FI (baseline) and MILP (proposed) methods, we consider another baseline variation where the supporter always stays on its default (lawn-mower) path and never explores to gather information outside this default path.
This experiment demonstrates the effectiveness of the exploration strategy in strategically gathering informative~data.

\subsection{Environment Setup}

\begin{figure*}
    \centering
    \subfigure[$32\times32$ Terrain Environment]{\includegraphics[trim = 0 0 0 5, clip, width = 0.25 \linewidth, height = 0.22 \linewidth]{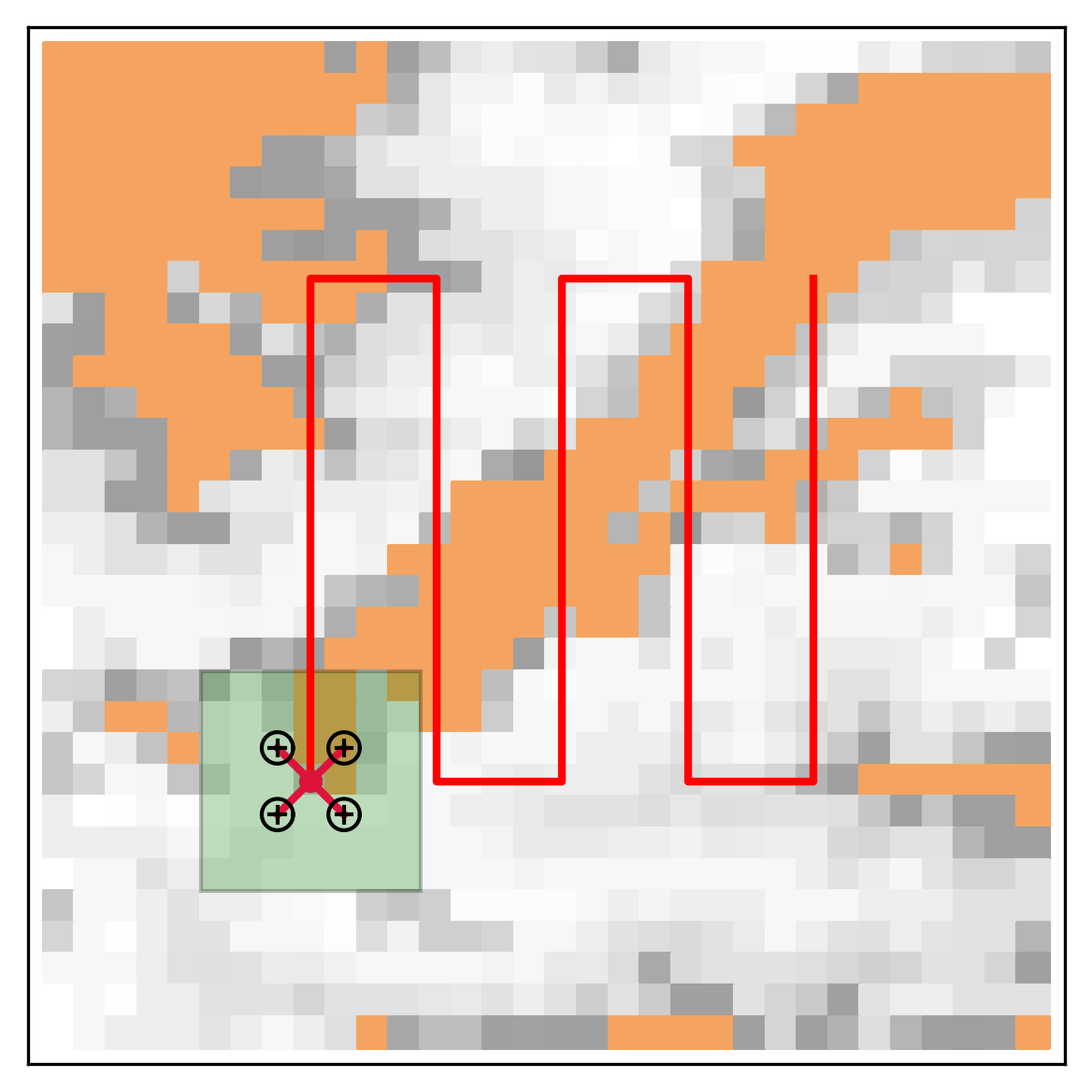}
    \label{Fig:6a}}
    \subfigure[Lawn Mower Exploration]{\includegraphics[trim = 0 0 0 5, clip, width = 0.33\linewidth]{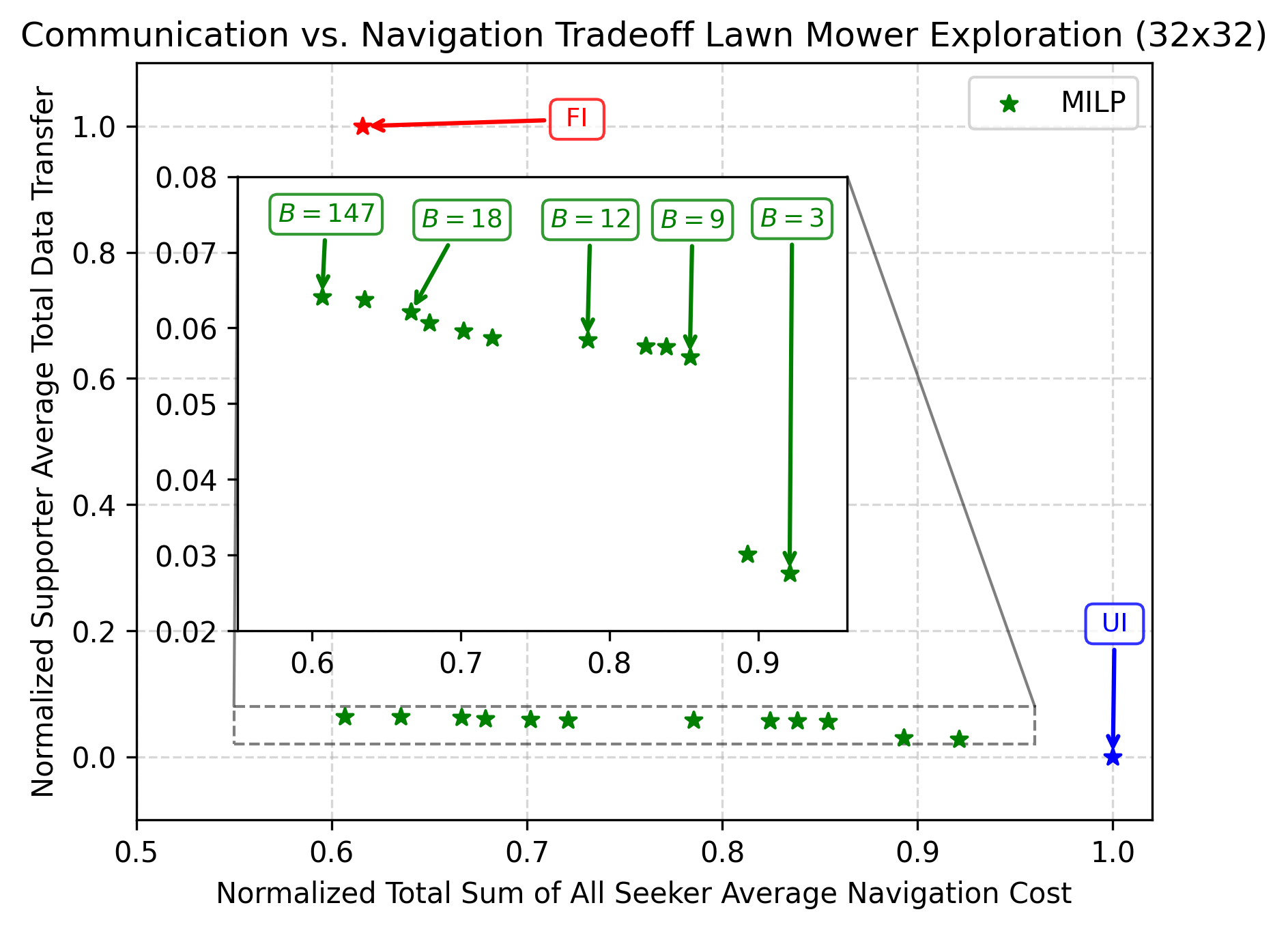}
    \label{Fig:6b}}
    \subfigure[Utility Based Exploration]{\includegraphics[trim = 0 0 0 5, clip, width = 0.33 \linewidth]{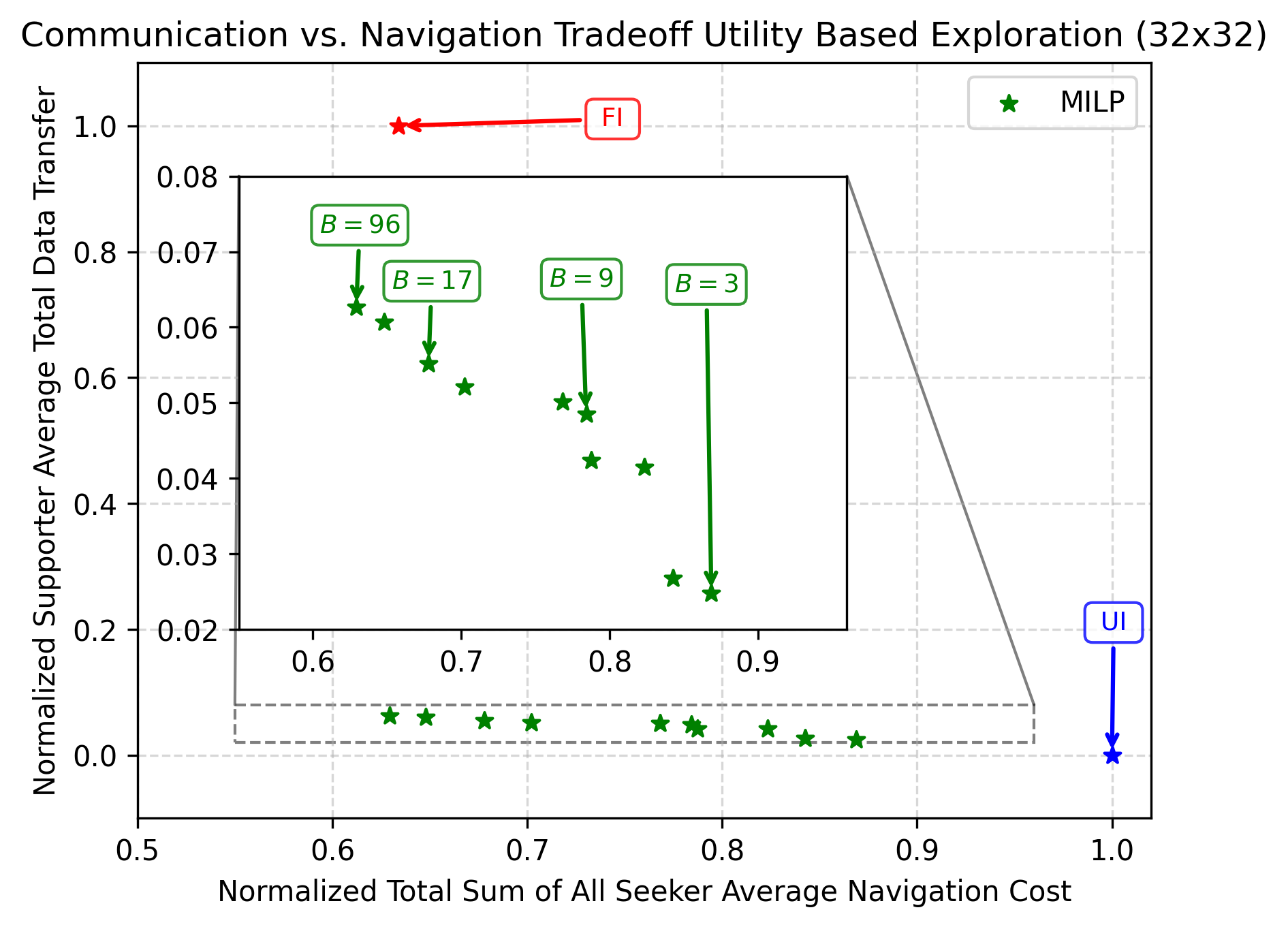}
    \label{Fig:6c}}
    \subfigure[$64\times64$ Terrain Environment]{\includegraphics[trim = 0 0 0 5, clip, width = 0.25 \linewidth, height = 0.22 \linewidth]{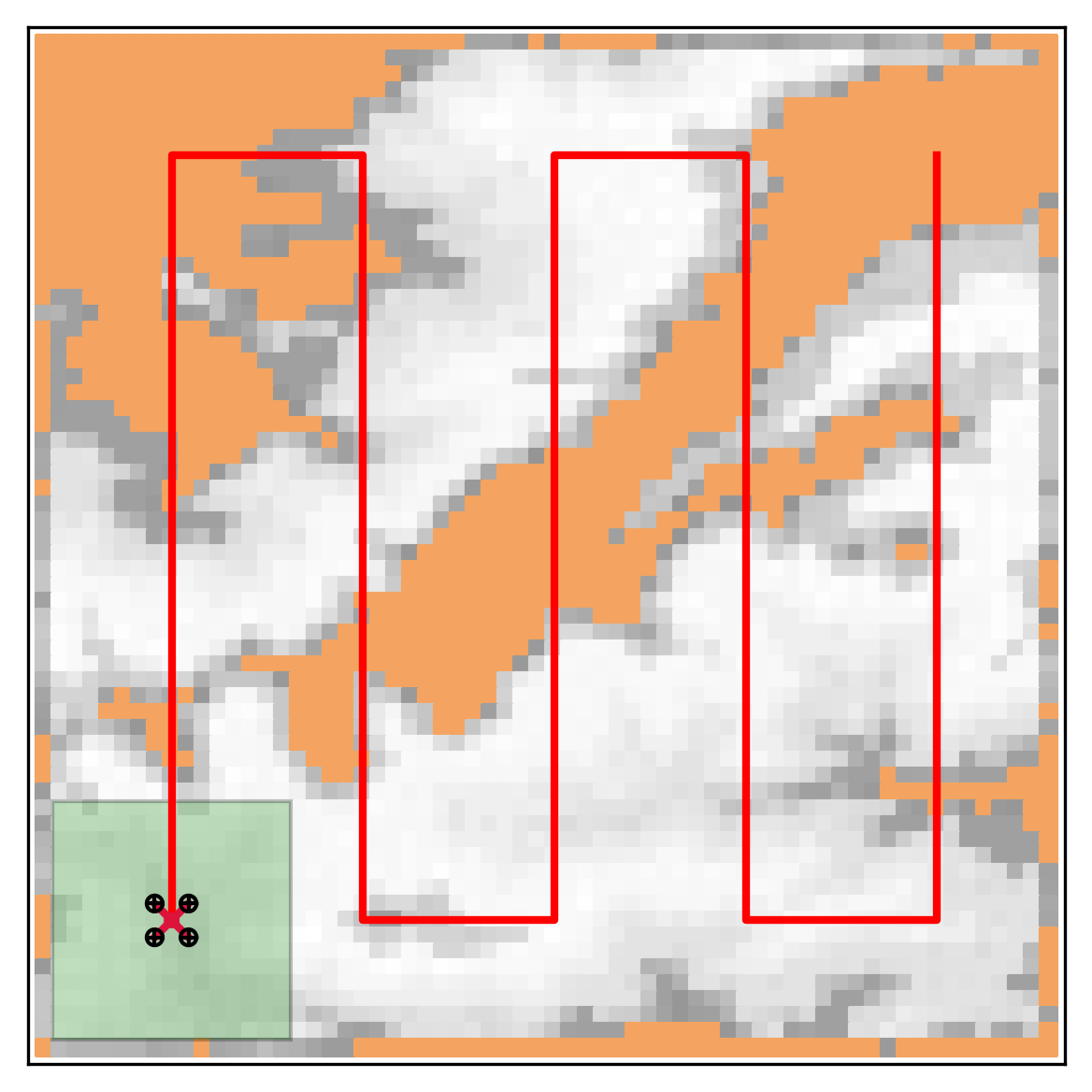}
    \label{Fig:6d}}
    \subfigure[Lawn Mower Exploration]{\includegraphics[trim = 0 0 0 5, clip, width = 0.33\linewidth]{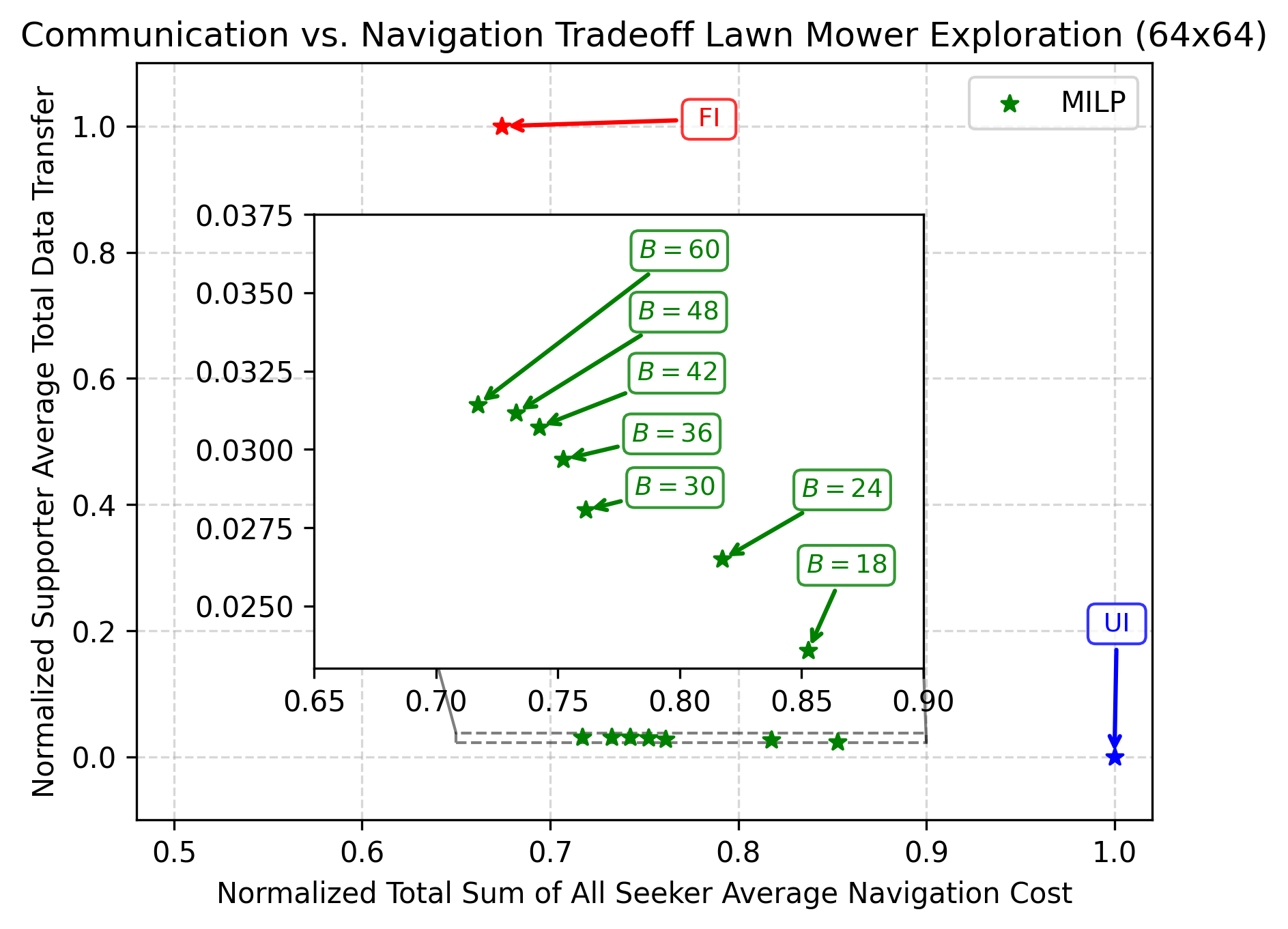}
    \label{Fig:6e}}
    \subfigure[Utility Based Exploration]{\includegraphics[trim = 0 0 0 5, clip, width = 0.33\linewidth]{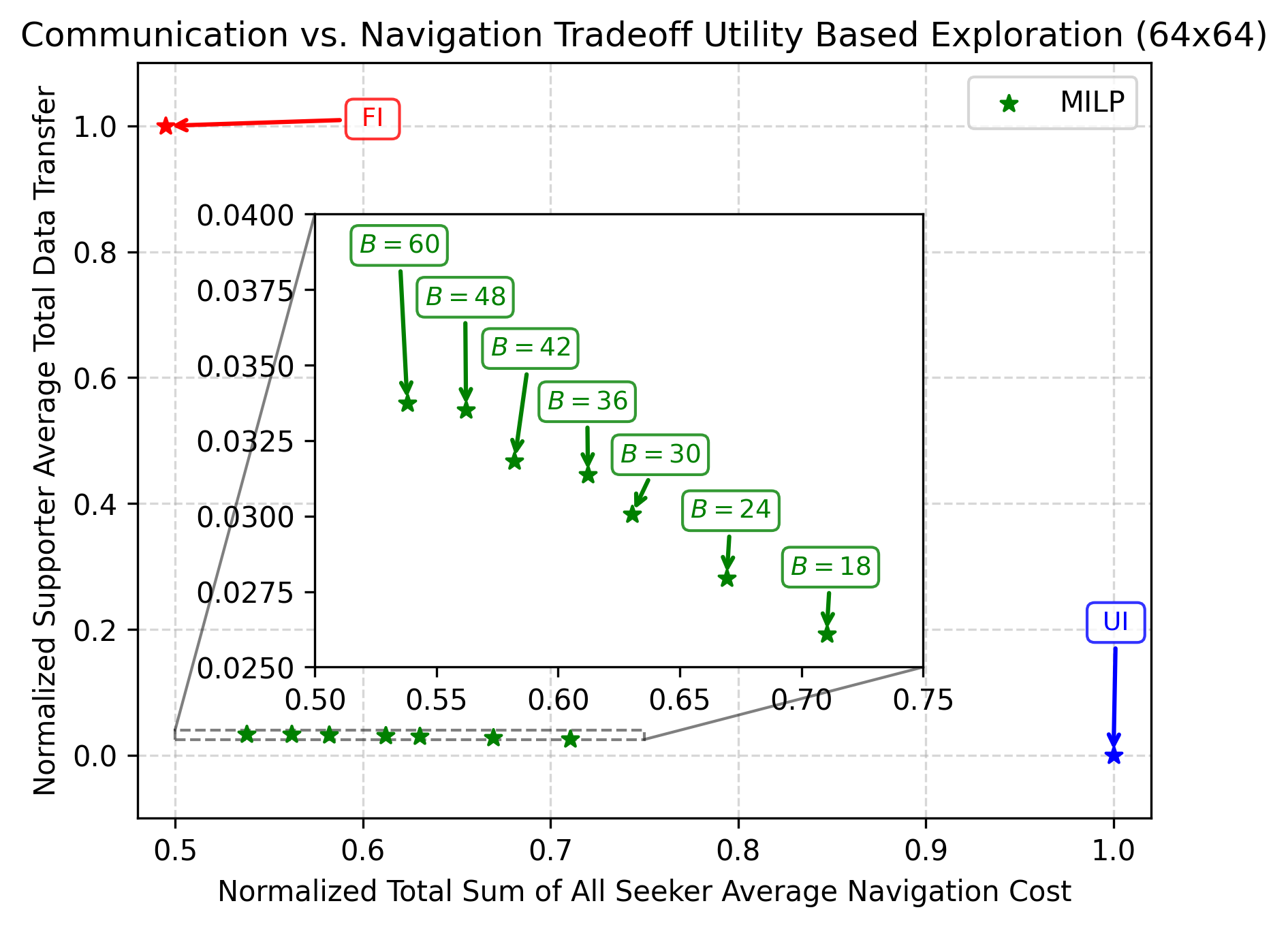}
    \label{Fig:6f}}
    
\caption{\small (a) Terrain map (32×32) with the supporter’s lawn-mower path (red) and its sensing area (green). (b)–(c) Normalized trade-off plots for lawn-mower and utility-based exploration on the 32×32 map. (d) Terrain map with high resolution (64×64) with the supporter’s lawn-mower path (red). (e)–(f) Normalized trade-off plots for the 64×64 map using both exploration methods.}
\label{Figure: 6} 
\vspace{-3mm}
\end{figure*}

\begin{figure*}
    \centering
    \subfigure[$30\times30$ Maze Environment]{\includegraphics[trim = 0 0 0 5, clip, width = 0.22 \linewidth, height = 0.22 \linewidth]{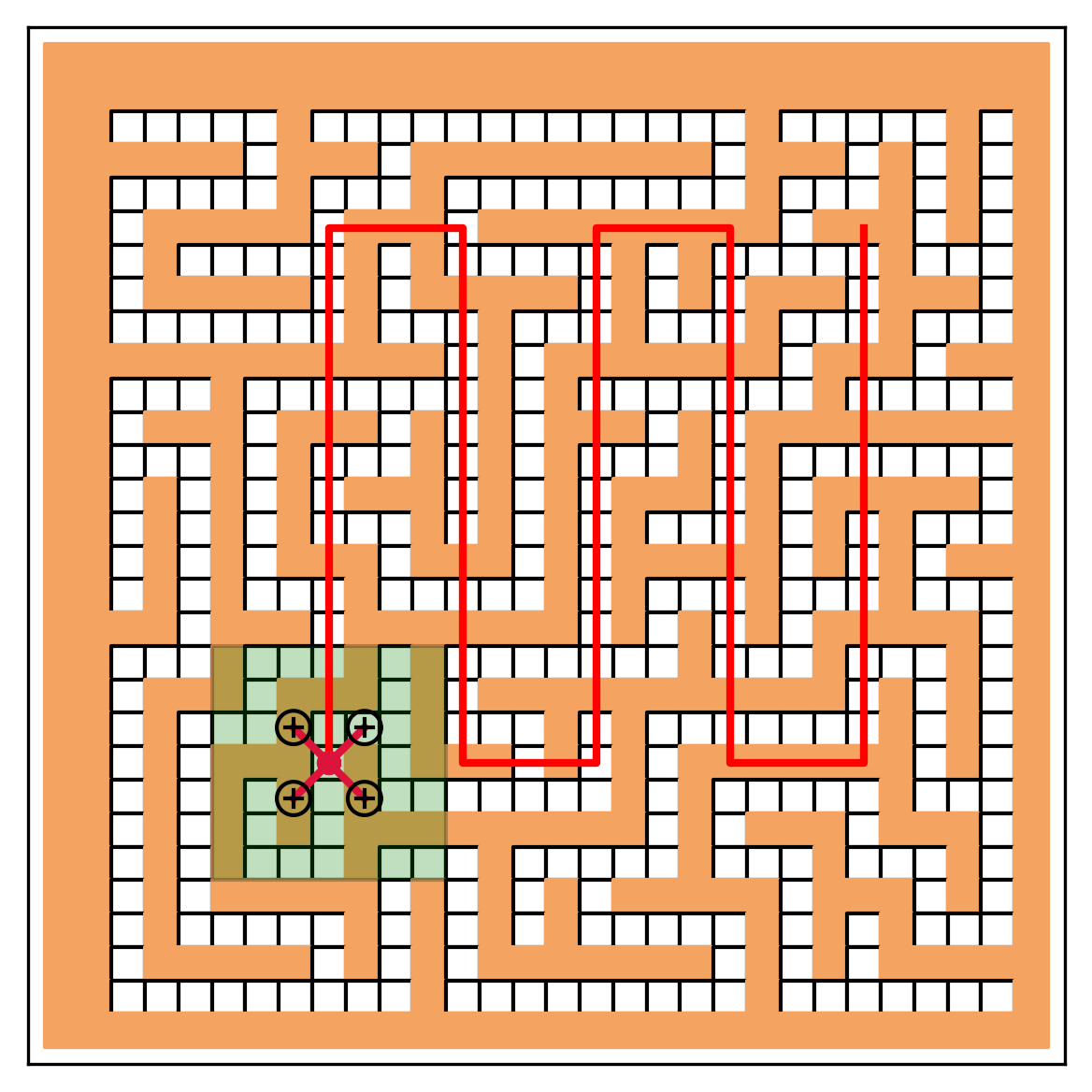}
    \label{Fig:7a}}
    \subfigure[Lawn Mower Exploration]{\includegraphics[trim = 0 0 0 5, clip, width = 0.33\linewidth]{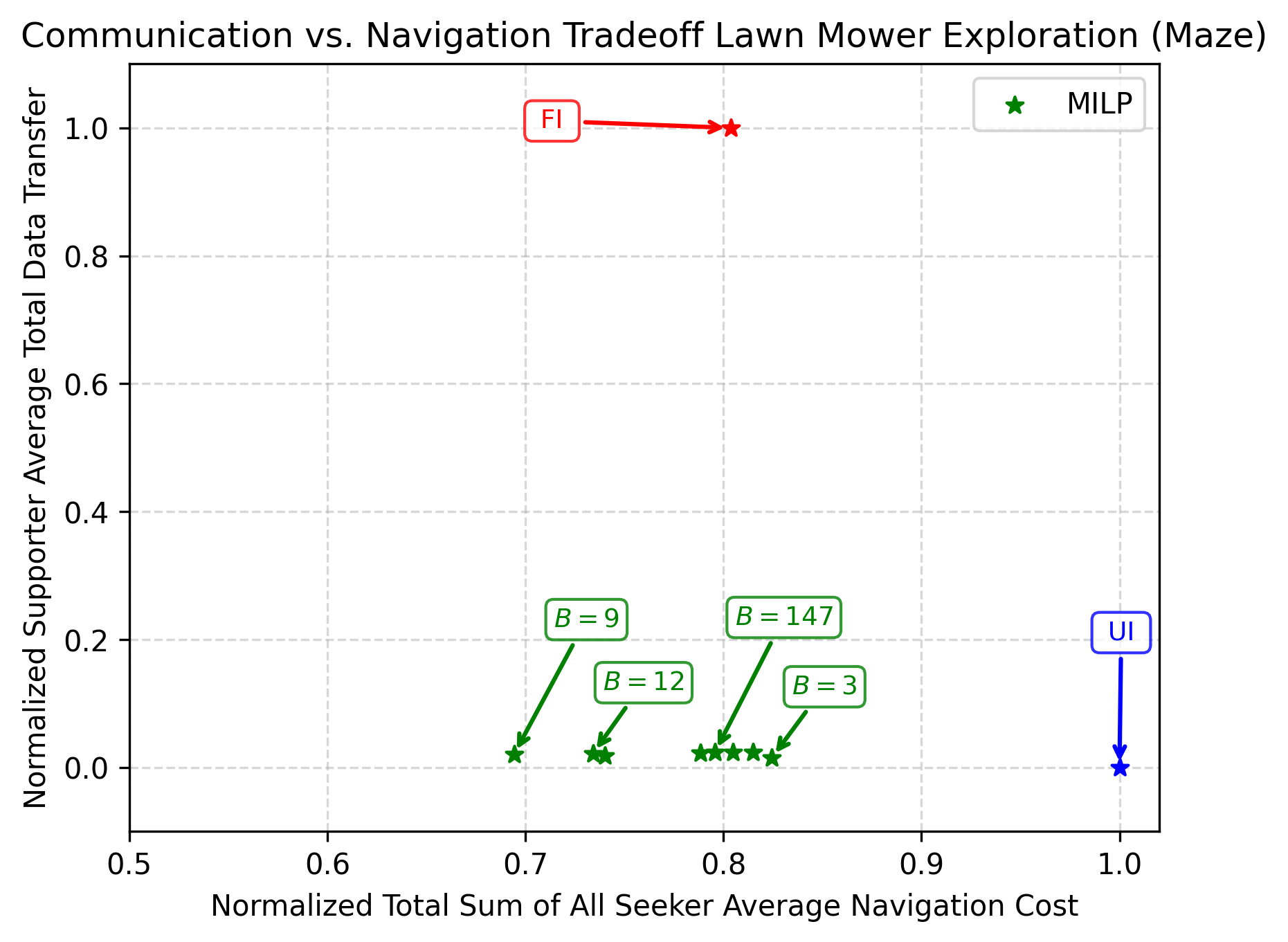}
    \label{Fig:7b}}
    \subfigure[Utility Based Exploration]{\includegraphics[trim = 0 0 0 5, clip, width = 0.33 \linewidth]{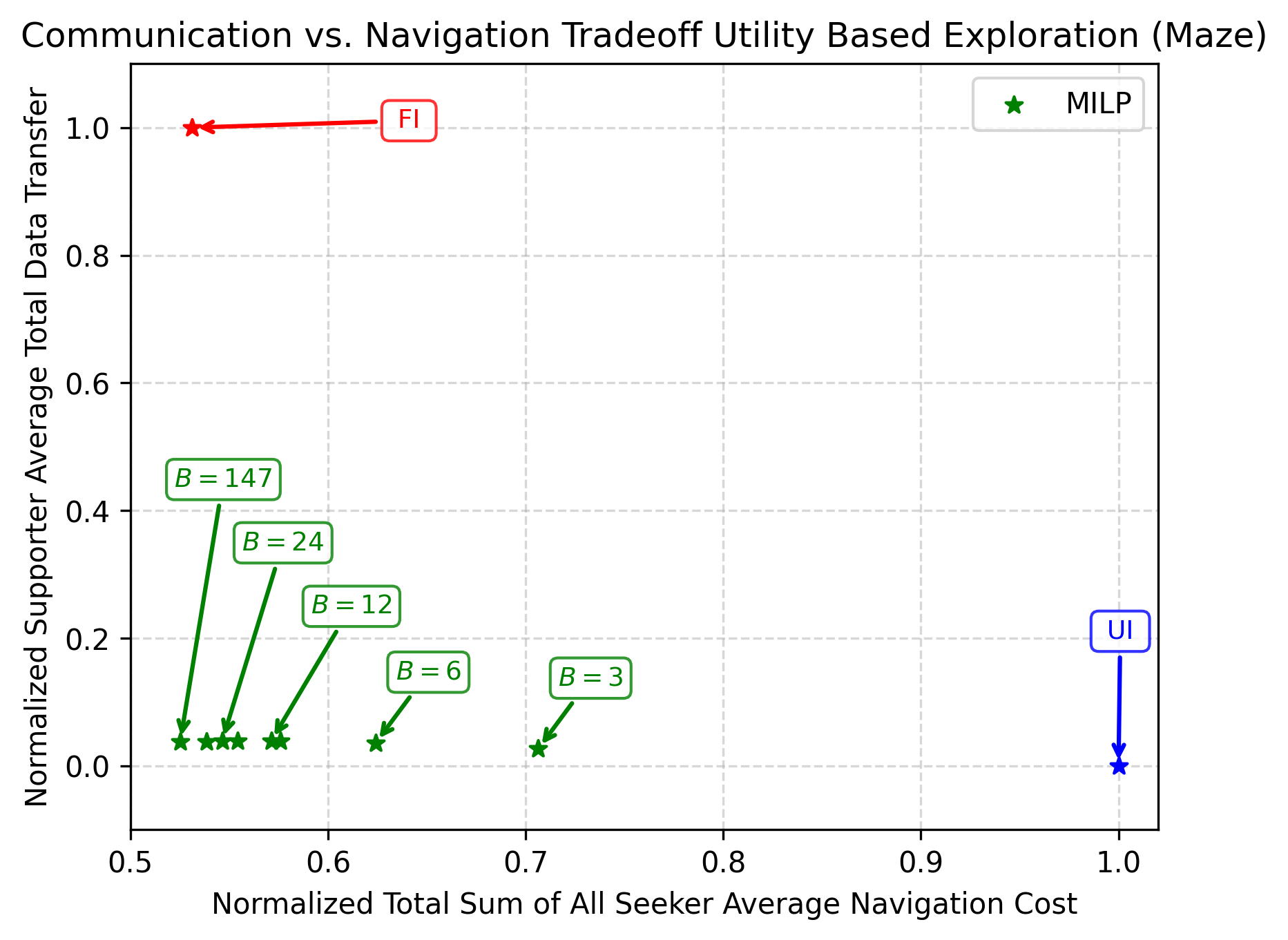}
    \label{Fig:7c}}
\caption{\small (a) Maze map with the supporter’s lawn-mower path (red) and sensing area (green). (b)-(c) Normalized trade-off plot for maze map with lawn mower and utility based exploration. Results of specific $B$ values highlighted with label for proposed \textit{MILP} method.}
\label{Figure: 7} 
\vspace{-6mm}
\end{figure*}

For the simulation experiments, two types of 2D environments are used: a terrain map (Figure~\ref{Fig:6a}) with varying occupancy levels for each cell and a maze map (Figure~\ref{Fig:7a}) with binary occupancy values $\{\phi^{\min}, \phi^{\max}\}$. 
To evaluate the scalability with environment sizes, two terrain environments were used with dimensions of $32 \times 32$ and $64 \times 64$, see Figures~\ref{Fig:6a}, \ref{Fig:6d} respectively, where shades of gray represent occupancy values (darker shades indicate higher occupancy). 
Both terrain maps represent the same environment, one with higher resolution than the other.
The maze environment has dimensions of $30 \times 30$, as shown in Figure~\ref{Fig:7a}.
For all cases, non-traversable cells are displayed in sandy-brown color. 
In all cases, the \textit{default path} of the supporter is chosen to be a lawn-mower path, as shown using the red lines in Figures~\ref{Fig:6a}, \ref{Fig:6d}, and \ref{Fig:7a}.
The supporter has a sensing window of $n^\h = 7$ for the maze and the $32\times 32$ terrain whereas $n^\s = 15$ for the $64 \times 64$ terrain map.

\subsection{Performance Metrics}

To evaluate the methods, we compare their performance over multiple random trials ($50$ trials) conducted on all the maps across all different settings discussed in the Baseline Methods section. 
To measure the team performance, we record the total amount of data transmitted by the supporter to the seekers, and the cumulative navigation cost for all seeker agents. 
The average total amount of data transmitted to all seeker agents over a series of simulations is given by 
\(\frac{1}{50} \sum_{n=1}^{50} \sum_{i=1}^{r} \mathcal{B}^{i,n}_{\mathbf{F}}\), where $\mathcal{B}^{i,n}_{\mathbf{F}}$ denotes the total amount of data sent to seeker $i$ at the $n$-th trial experiment, under the algorithm $\mathbf{F} \in $\{UI, FI$_0$, MILP$_0$, FI$_1$, and MILP$_1$\}. 
Here FI$_0$ represents the FI method when the supporter always stays on its default path and FI$_1$ denotes the case when the supporter follows our proposed strategic exploration strategy in \Cref{subsec:exploration}.
 MILP$_0$ and MILP$_1$ are defined analogously.
Since no data is transferred in the UI method, the exploration strategy of the support does not affect the performance. 
Finally, the average navigation cost of all seeker agents is computed as 
\(\frac{1}{50} \sum_{n=1}^{50} \sum_{i=1}^{r} \mathcal{C}^{i,n}_{\mathbf{F}}\).

\subsection{Simulation Results}

The trade-off plots in Figures~\ref{Figure: 6}-\ref{Figure: 7} show the simulation results for a team of three seekers on both the terrain and maze maps. 
Each trade-off plot has its $y$-axis normalized by the total communication amount of the \text{FI} method and its $x$-axis normalized by the navigation cost of the \text{UI} method.
Each green star in these figures is obtained by choosing a different bandwidth limit. 
The collection of these green stars illustrates the communication-navigation trade-off curve (i.e., the pareto frontier) of our method.

Figures~\ref{Fig:6b}--\ref{Fig:6c} illustrate the comparison between the supporter's lawn-mower and utility-based exploration strategies on the $32\times32$ terrain map. 
It can be observed that, under utility-based exploration, the average total navigation cost incurred by all seeker agents is reduced. 
Furthermore, the amount of data transmitted is significantly lower when using the proposed VoI-based \text{MILP} method across a range of bandwidth values. 
As the value of \(B\) increases, data transmission from the supporter gradually increases and the total navigation cost of all seekers is reduced. 
%
Figures~\ref{Fig:6e}-\ref{Fig:6f} demonstrate the same artifact for the higher resolution map environment ($64\times64$).
We notice the benefit of our method is more prominent in larger environments (i.e., when seekers know very little about the environment).

In summary, the main observations are as follows:
\begin{itemize}
    \item Significantly less communication overhead compared to FI across all instances.
    \item Significant improvement in navigation cost over UI across all cases.
    \item Utility-based exploration lowers navigation cost.
    \item Natural communication-navigation trade-off as bandwidth is varied.
\end{itemize}

Similarly, Figures~\ref{Fig:7b}--\ref{Fig:7c} show the trade-off plots for the lawn-mower and utility-based exploration methods in the maze environment, also for a team of three seekers. An additional key observation was made in this experiment:

\paragraph{More information is not always beneficial}

The trade-off plot in Figure~\ref{Fig:7b} shows that, for certain $B$ values, the average navigation cost for seekers is lower with the MILP$_0$ method than for higher $B$ values and the FI$_0$ method, where the supporter transmits new observations instantaneously. 
This occurs because the lawn-mower pattern leads the supporter to explore non-relevant areas and transmit more of those observations.
These non-relevant transmissions cause seekers to explore unnecessarily and take longer paths to their goals. 
This behavior arises from the small maze environment used in the experiment, where subtle twists and turns can cause deviations.
A similar trend is observed in~\cite{lessismore}; although that work examines dense communication links rather than high-volume information exchange, it reaches a comparable conclusion that excessive communication can hinder adaptation.

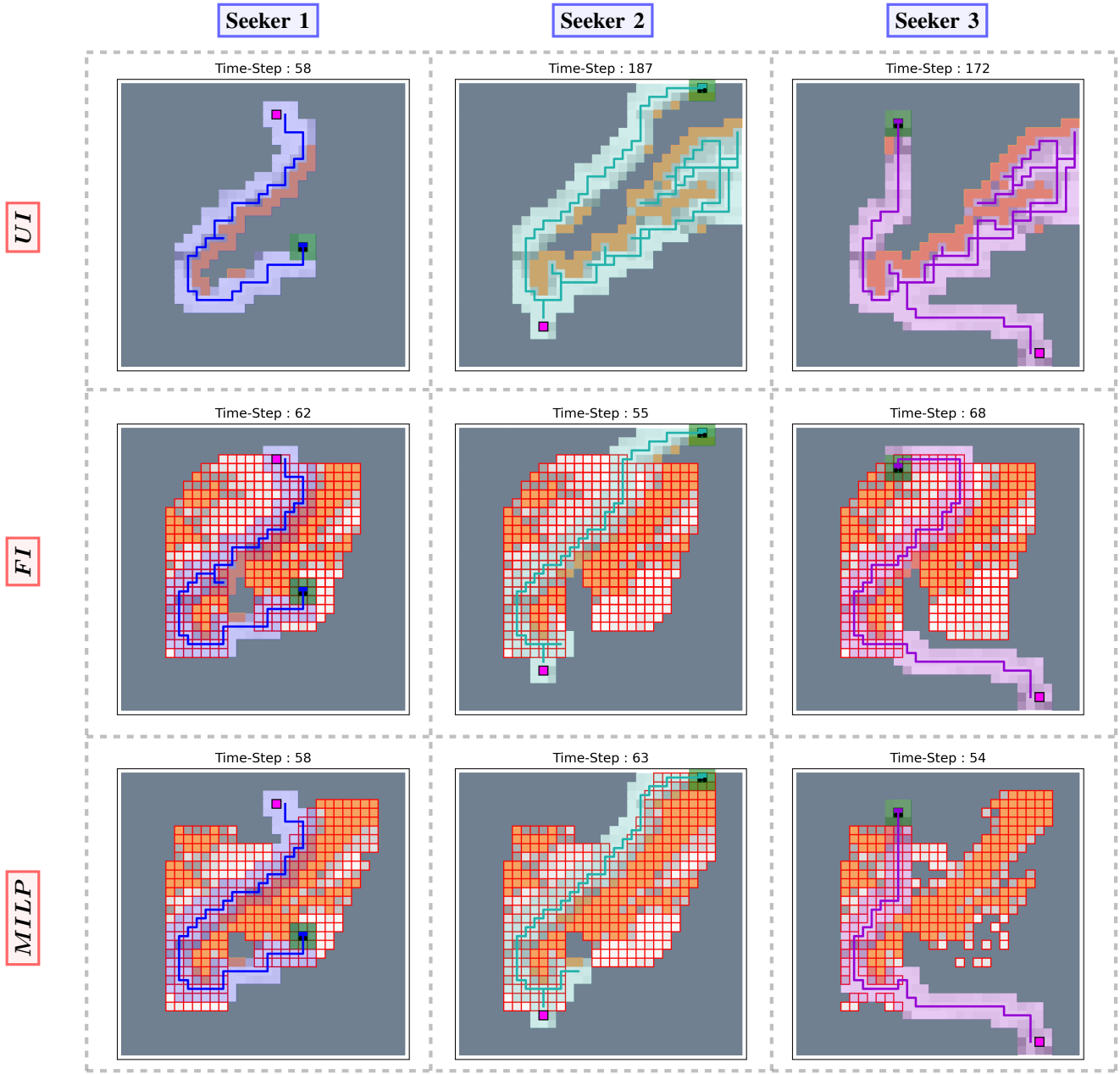
\begin{figure*}[h]
    \vspace{0.5mm}
    \centering
    \hspace{10mm}
    \subfigure{
    \includegraphics[width=0.27\linewidth]{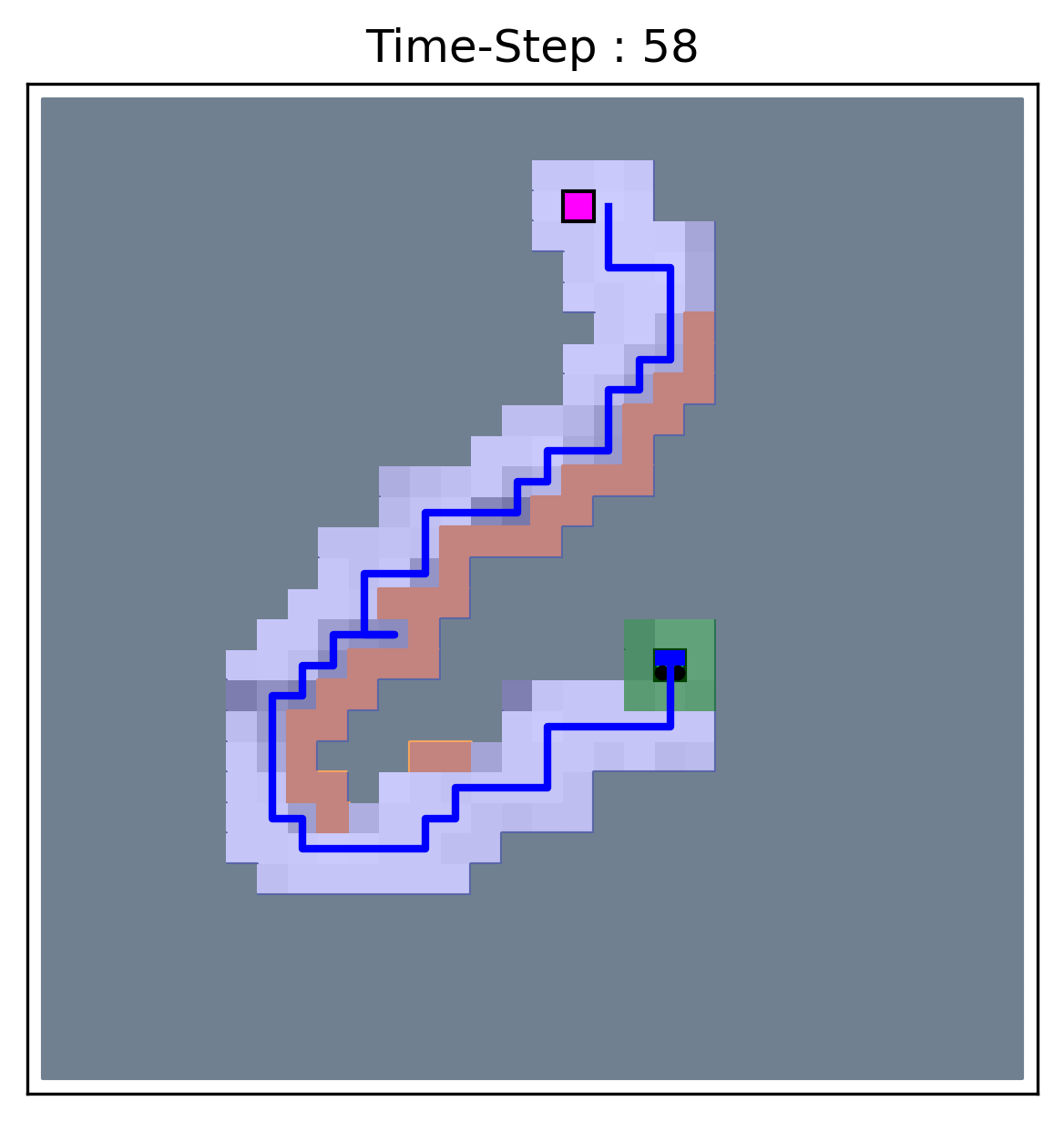}
    \label{Fig:8a}
    }
    \subfigure{
    \includegraphics[width=0.27\linewidth]{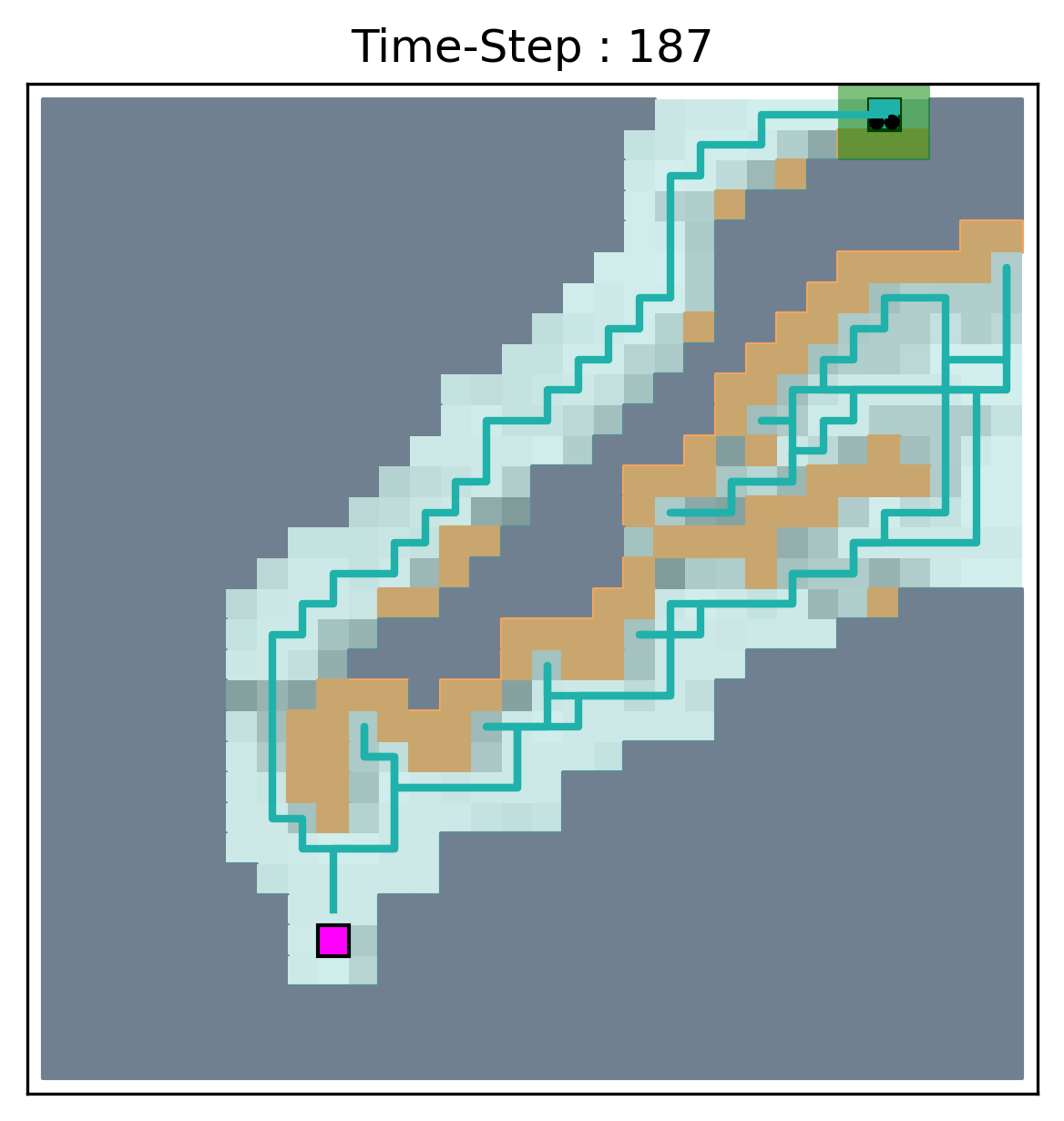}
    \label{Fig:8b}
    }
    \subfigure{
    \includegraphics[width=0.27\linewidth]{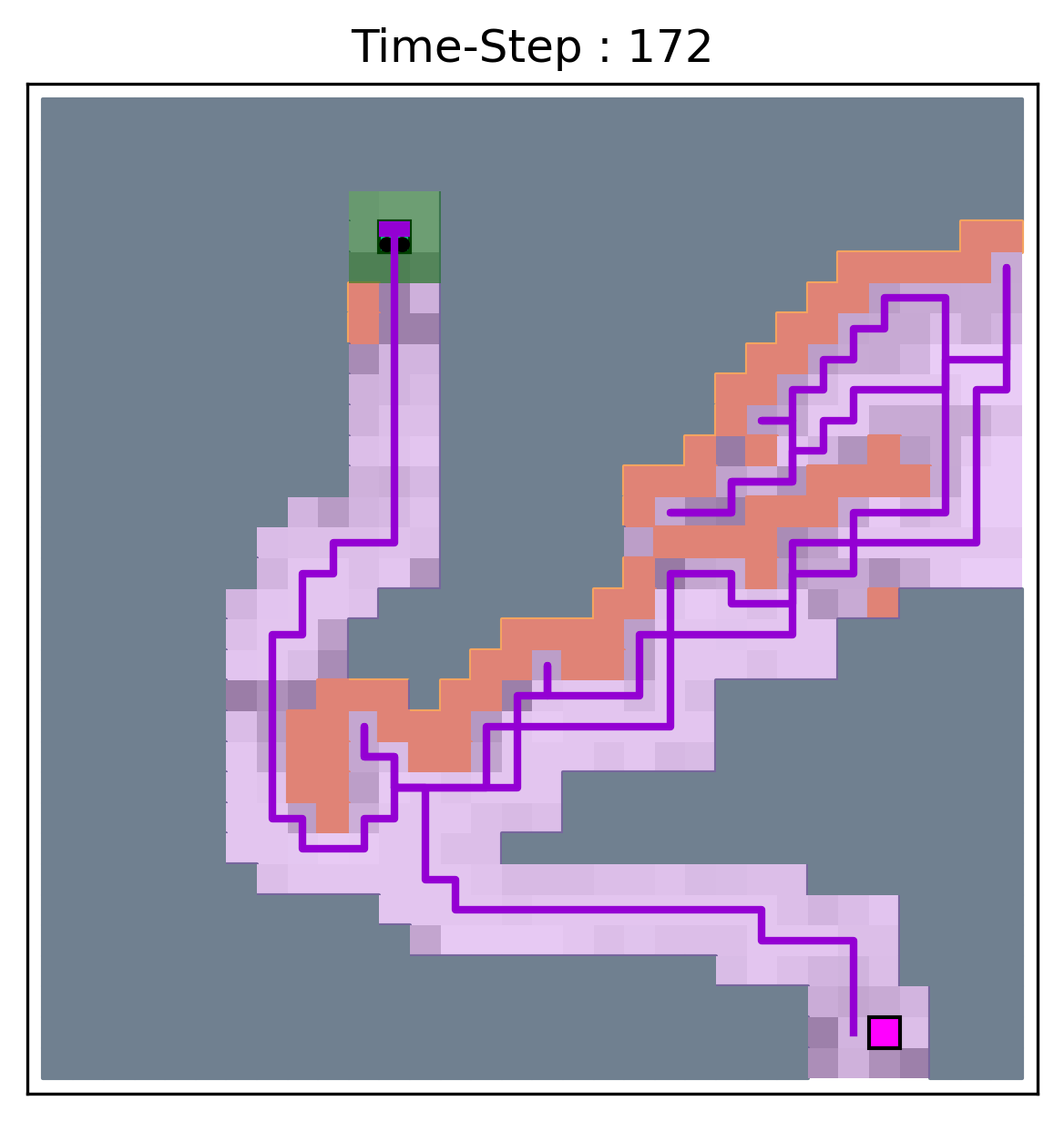}
    \label{Fig:8c}
    }
    \par\hspace{10mm}
    \subfigure{
    \includegraphics[width=0.27\linewidth]{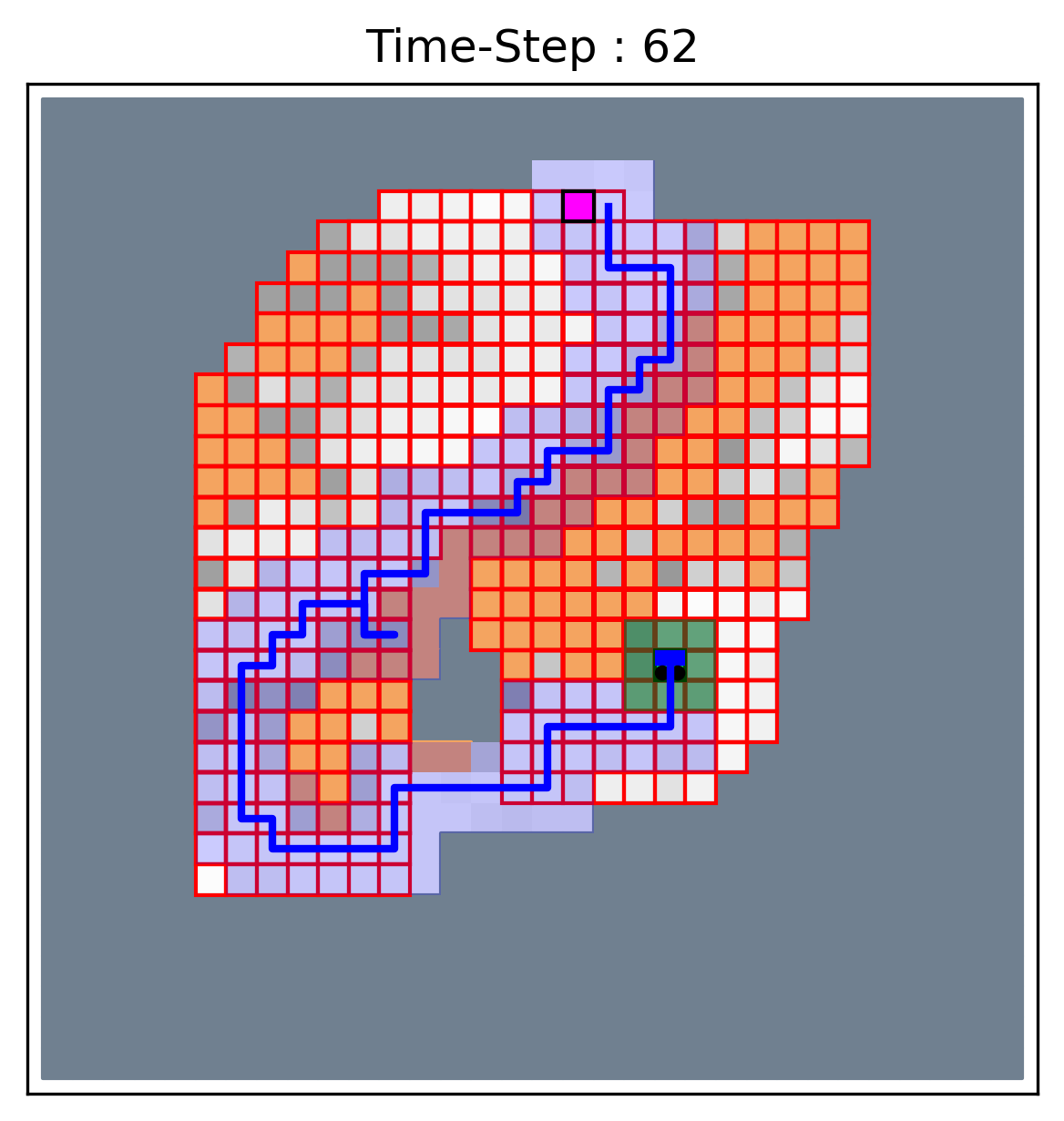}
    \label{Fig:8d}
    }
    \subfigure{
    \includegraphics[width=0.27\linewidth]{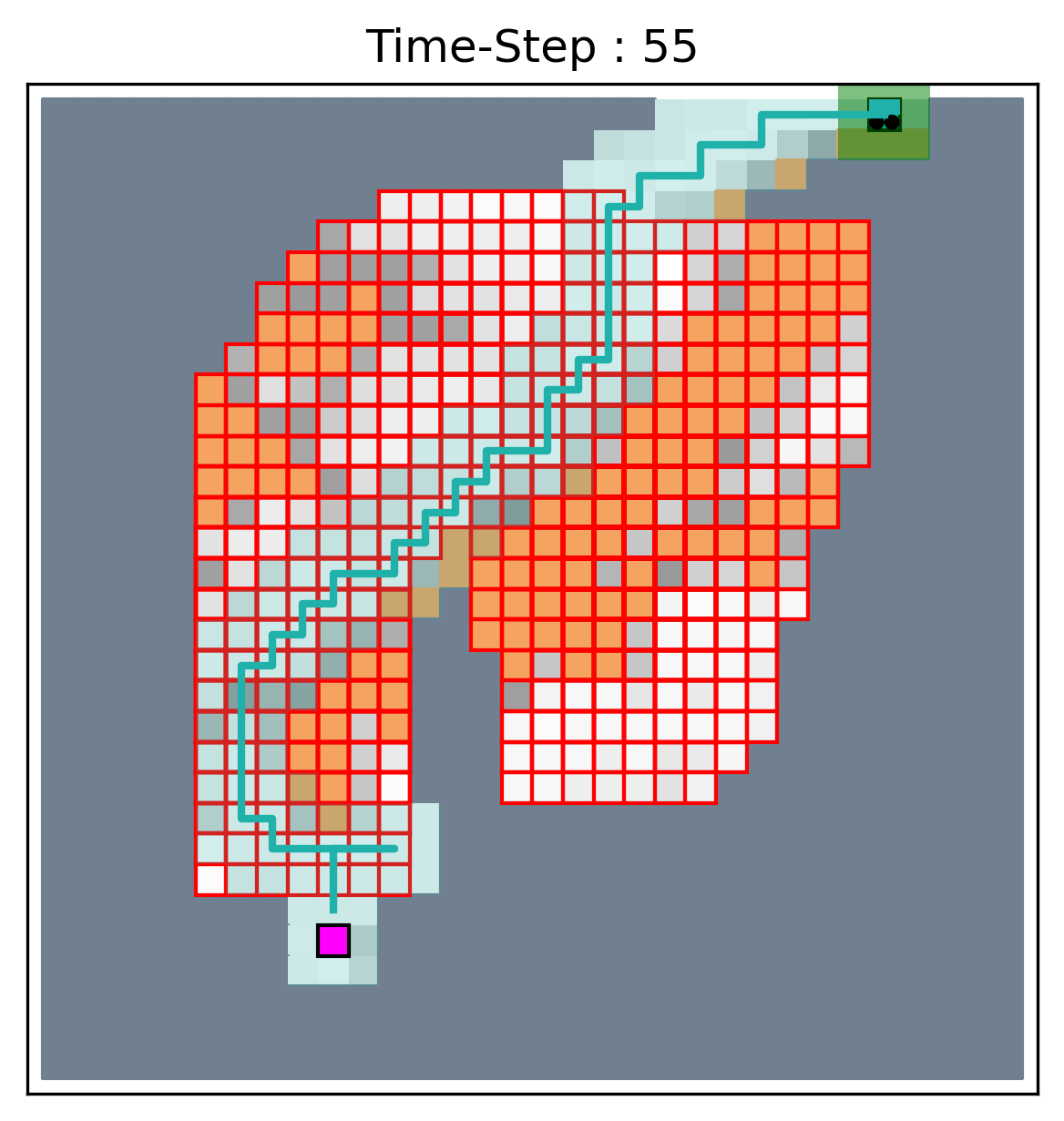}
    \label{Fig:8e}
    }
    \subfigure{
    \includegraphics[width=0.27\linewidth]{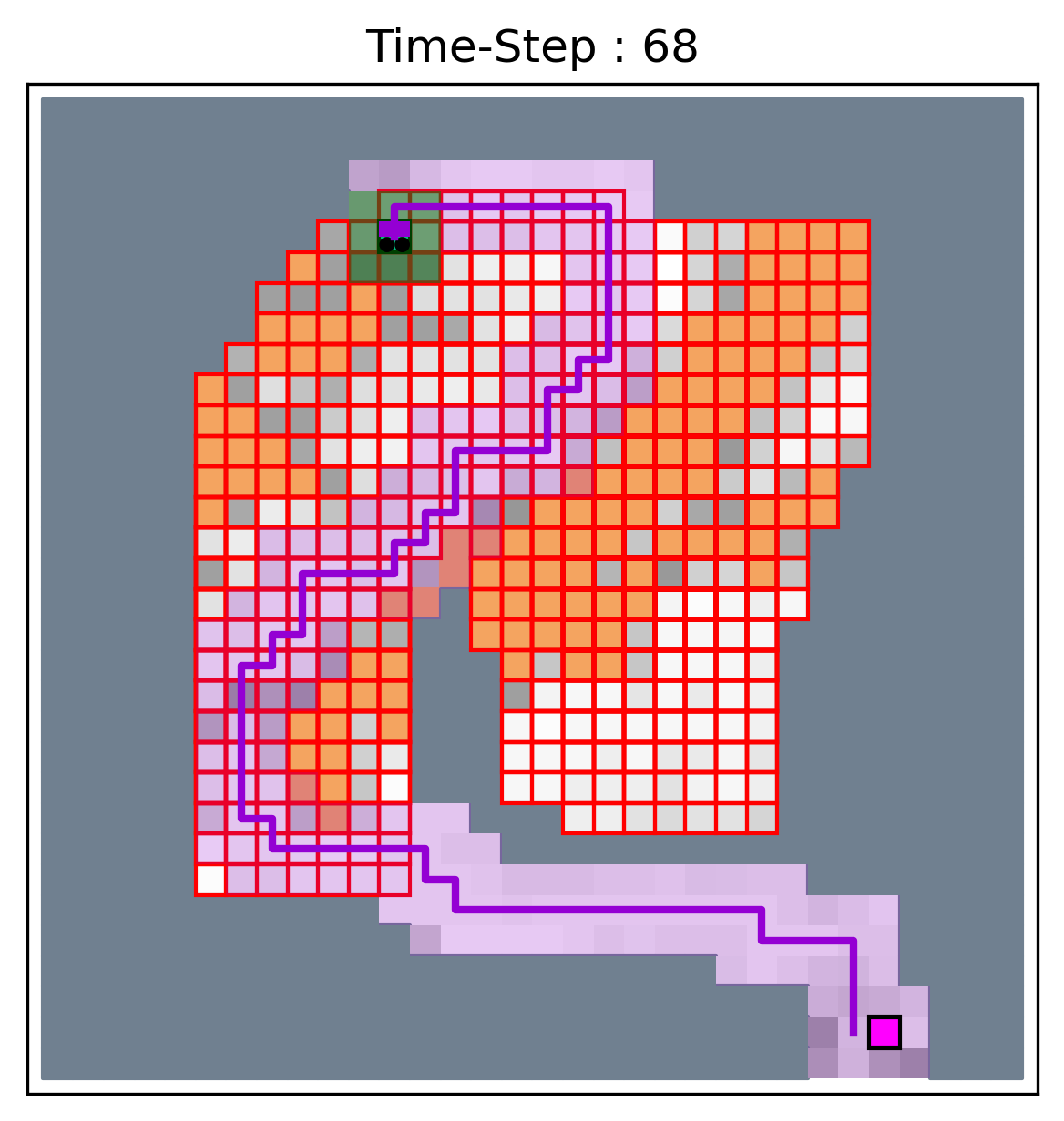}
    \label{Fig:8f}
    }
    \par\hspace{10mm}
    \subfigure{
    \includegraphics[width=0.27\linewidth]{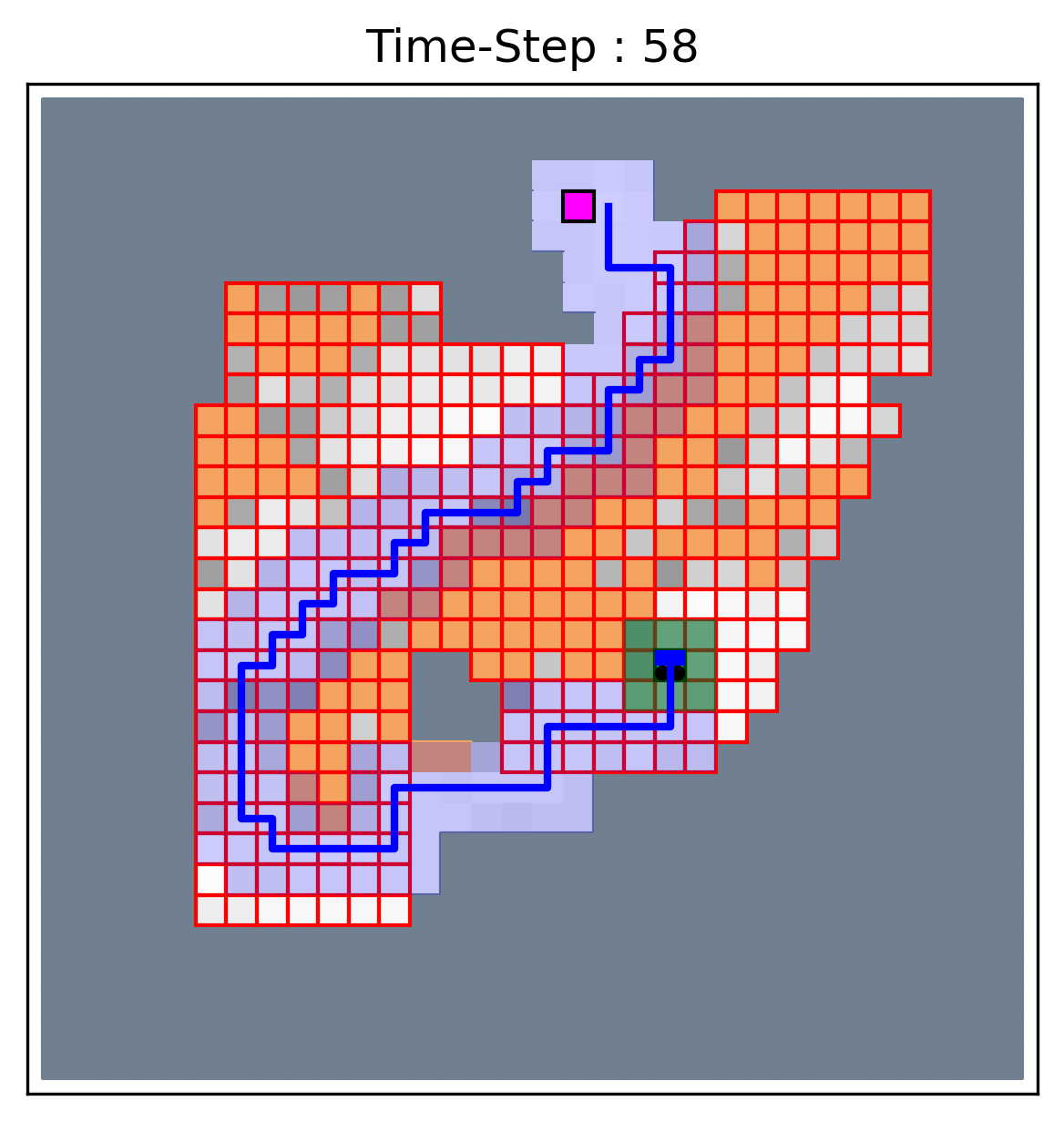}
    \label{Fig:8g}
    }
    \subfigure{
    \includegraphics[width=0.27\linewidth]{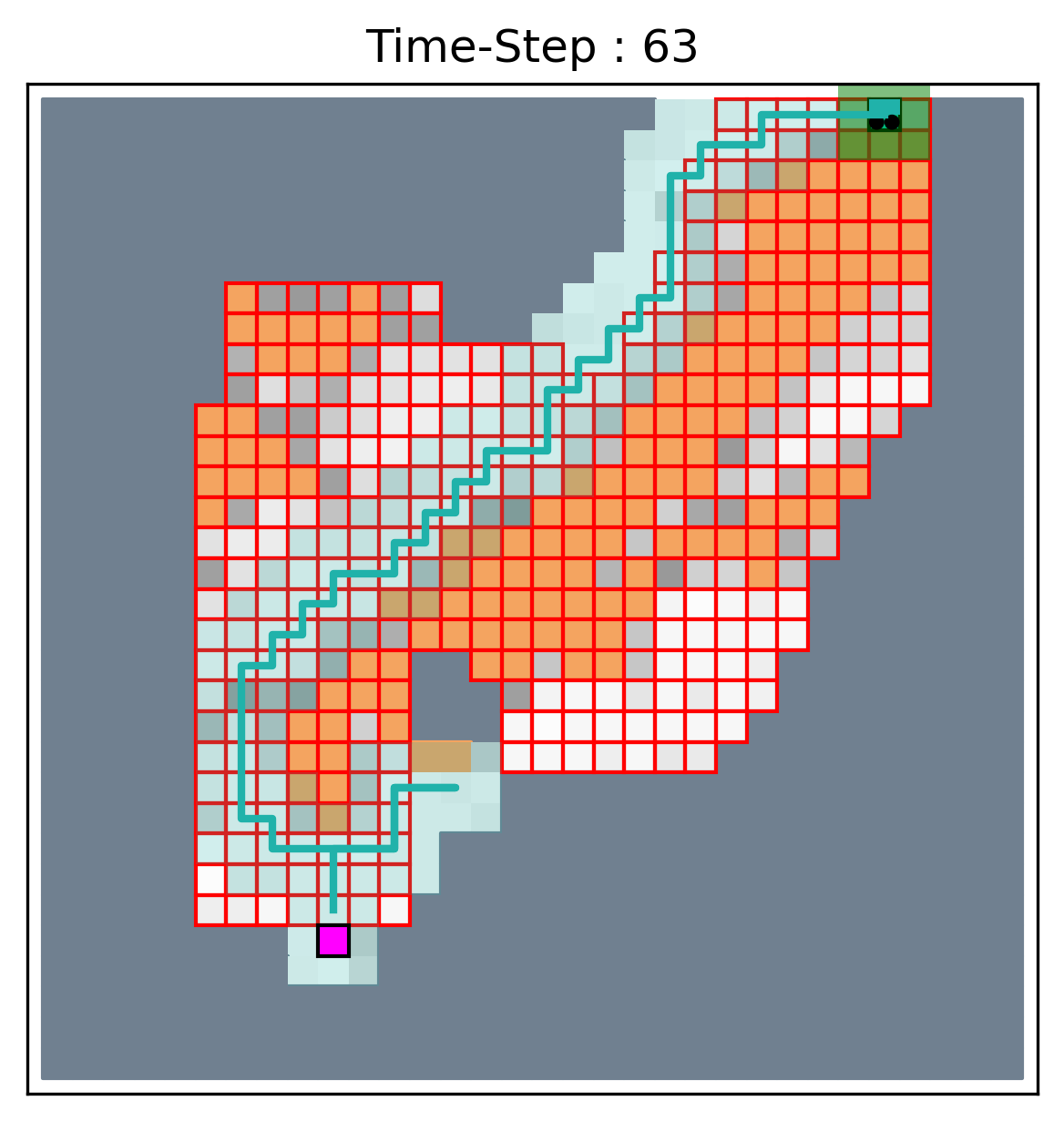}
    \label{Fig:8h}
    }
    \subfigure{
    \includegraphics[width=0.27\linewidth]{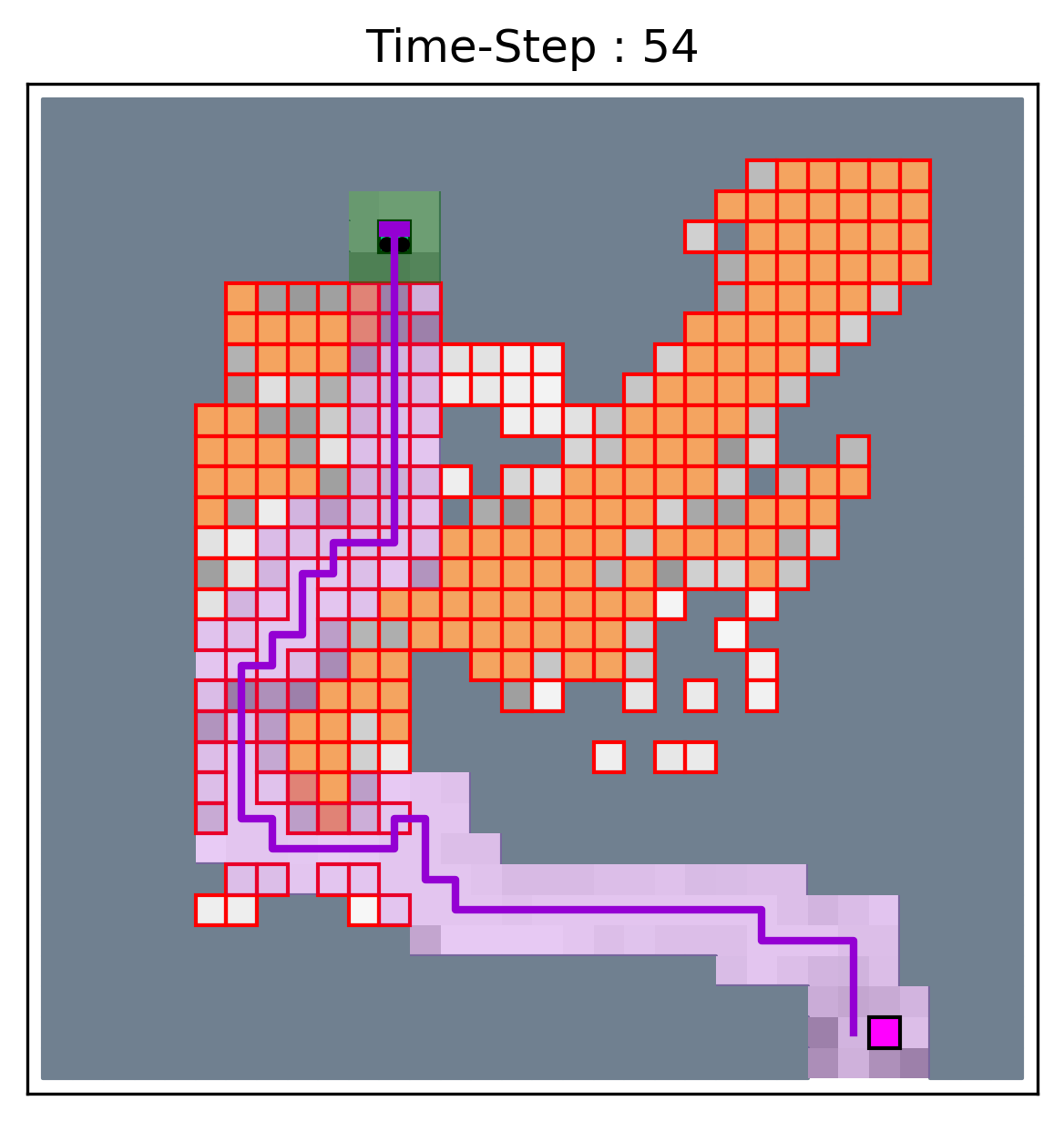}
    \label{Fig:8i}
    }
    \caption{\small Simulation frames of a three-seeker–one-supporter team operating on a $32\times32$ terrain. Each column shows a seeker’s final exploration map at its goal location, with the team of seekers shown in blue, lightseagreen, and dark-violet for the first, second, and third columns, respectively. The regions explored by each seeker are highlighted with its corresponding transparent color shade, while the supporter-transmitted cell locations up to each seeker reaching its final goal are marked with red edges. The magenta cell indicates the seeker’s start location, and the final navigation path is shown as a line in its respective color. The supporter performs utility-based exploration for the team. The rows represent different communication strategies: uninformed (UI) exploration in the first row, fully informed (FI$_1$) exploration in the second, and the proposed VoI- and MILP$_1$-based method in the third, with maximum allowed bandwidth $B = 27$ for the MILP$_1$ case.}
    
    \begin{tikzpicture}[remember picture, overlay,
    rowtitlenode/.style={rectangle, draw=red!60, fill=red!5, very thick, minimum size=5mm},
    coltitlenode/.style={rectangle, draw=blue!60, fill=blue!5, very thick, minimum size=5mm}]
        \begin{scope}[shift={(0.0,2.5)}]
            \draw[gray, ultra thick, dashed, opacity = 0.5] (-7.5cm, 1.0cm) -- (-7.5cm, 17.0cm); 
            \draw[gray, ultra thick, dashed, opacity = 0.5] (-2.1cm, 1.0cm) -- (-2.1cm,17.0cm); 
            \draw[gray, ultra thick, dashed, opacity = 0.5] (3.25cm, 1.0cm) -- (3.25cm, 17.0cm);
            \draw[gray, ultra thick, dashed, opacity = 0.5] (8.65cm, 1.0cm) -- (8.65cm, 17.0cm);
            \draw[gray, ultra thick, dashed, opacity = 0.5] (-7.5cm, 1.0cm) -- (8.65cm, 1.0cm);
            \draw[gray, ultra thick, dashed, opacity = 0.5] (-7.5cm, 6.25cm) -- (8.65cm, 6.25cm); 
            \draw[gray, ultra thick, dashed, opacity = 0.5] (-7.5cm, 11.70cm) -- (8.65cm, 11.70cm); 
            \draw[gray, ultra thick, dashed, opacity = 0.5] (-7.5cm, 17.0cm) -- (8.65cm, 17.0cm);
            \node[rowtitlenode,rotate=90] at (-8.5cm, 14.25cm) {$\boldsymbol{UI}$};
            \node[rowtitlenode,rotate=90] at (-8.5cm, 9.0cm) {$\boldsymbol{FI}$};
            \node[rowtitlenode,rotate=90] at (-8.5cm, 3.4cm) {$\boldsymbol{MILP}$};
            \node[coltitlenode] at (-4.65cm, 17.5cm) {\textbf{Seeker 1}};
            \node[coltitlenode] at (0.60cm, 17.5cm) {\textbf{Seeker 2}};
            \node[coltitlenode] at (5.85cm, 17.5cm) {\textbf{Seeker 3}};
        \end{scope}
    \end{tikzpicture}
    \label{Figure: 8}
\end{figure*}

Figure~\ref{Figure: 8} shows the simulation frames of all seekers' final exploration maps across the proposed and baseline methods for the $32\times32$ terrain environment.
For this simulation, the seekers’ start locations are $\mathbf{S}^1 = (17,28)$, $\mathbf{S}^2 = (9,4)$, and $\mathbf{S}^3 = (27,1)$, and their goal locations are $\mathbf{G}^1 = (20,13)$, $\mathbf{G}^2 = (27,31)$, and $\mathbf{G}^3 = (11,27)$, respectively.
The supporter starts at $\mathbf{S}^\h = (8,8)$ and performs utility-based exploration to assist the team of seekers. The maximum allowable bandwidth used by the supporter in the proposed VoI- and MILP-based approach is $B=27$.

\begin{figure*}
    \vspace{0.25mm}
    \centering
    \hspace{10mm}
    \subfigure{
    \includegraphics[width=0.27\linewidth]{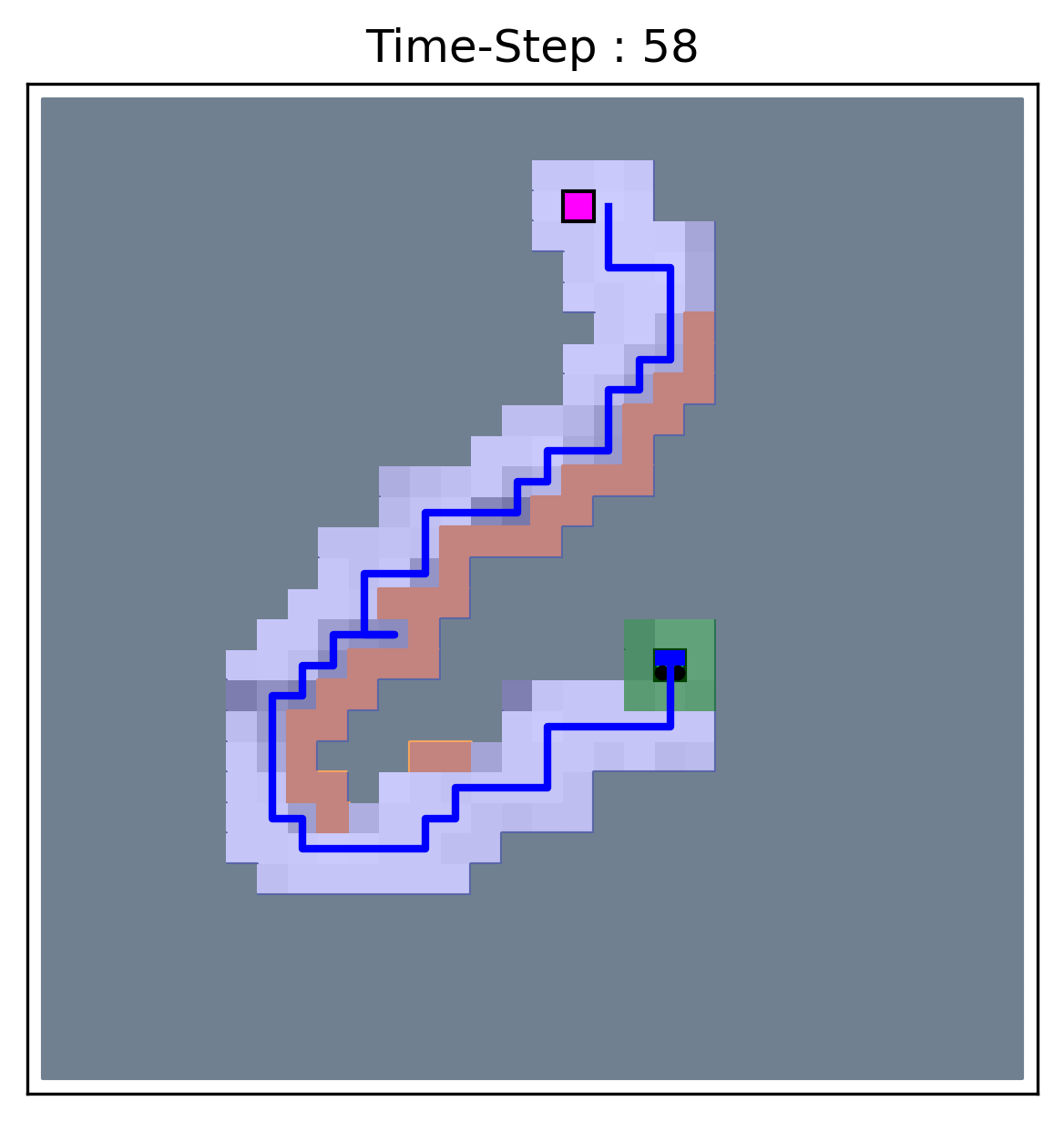}
    \label{Fig:9a}
    }
    \subfigure{
    \includegraphics[width=0.27\linewidth]{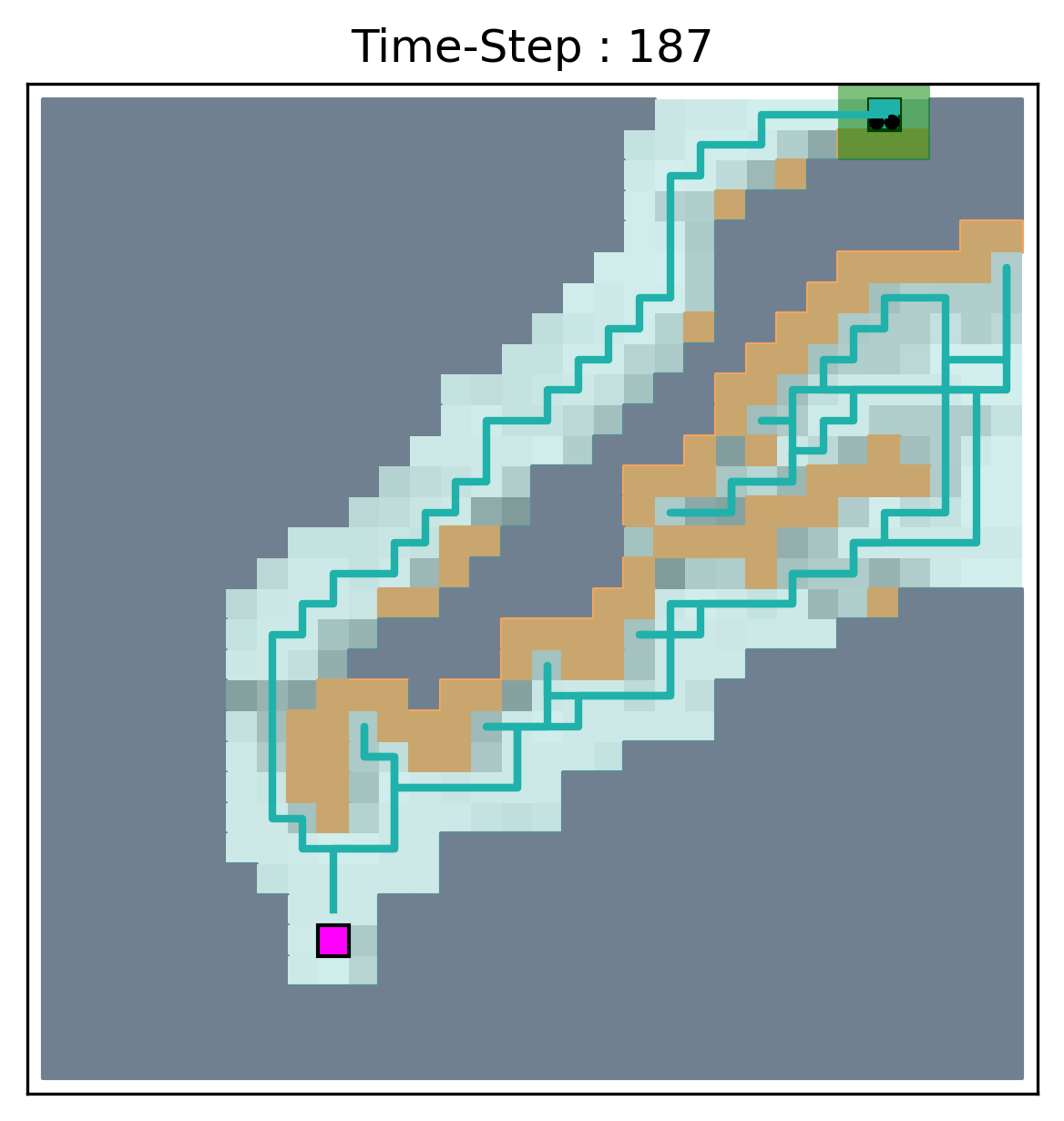}
    \label{Fig:9b}
    }
    \subfigure{
    \includegraphics[width=0.27\linewidth]{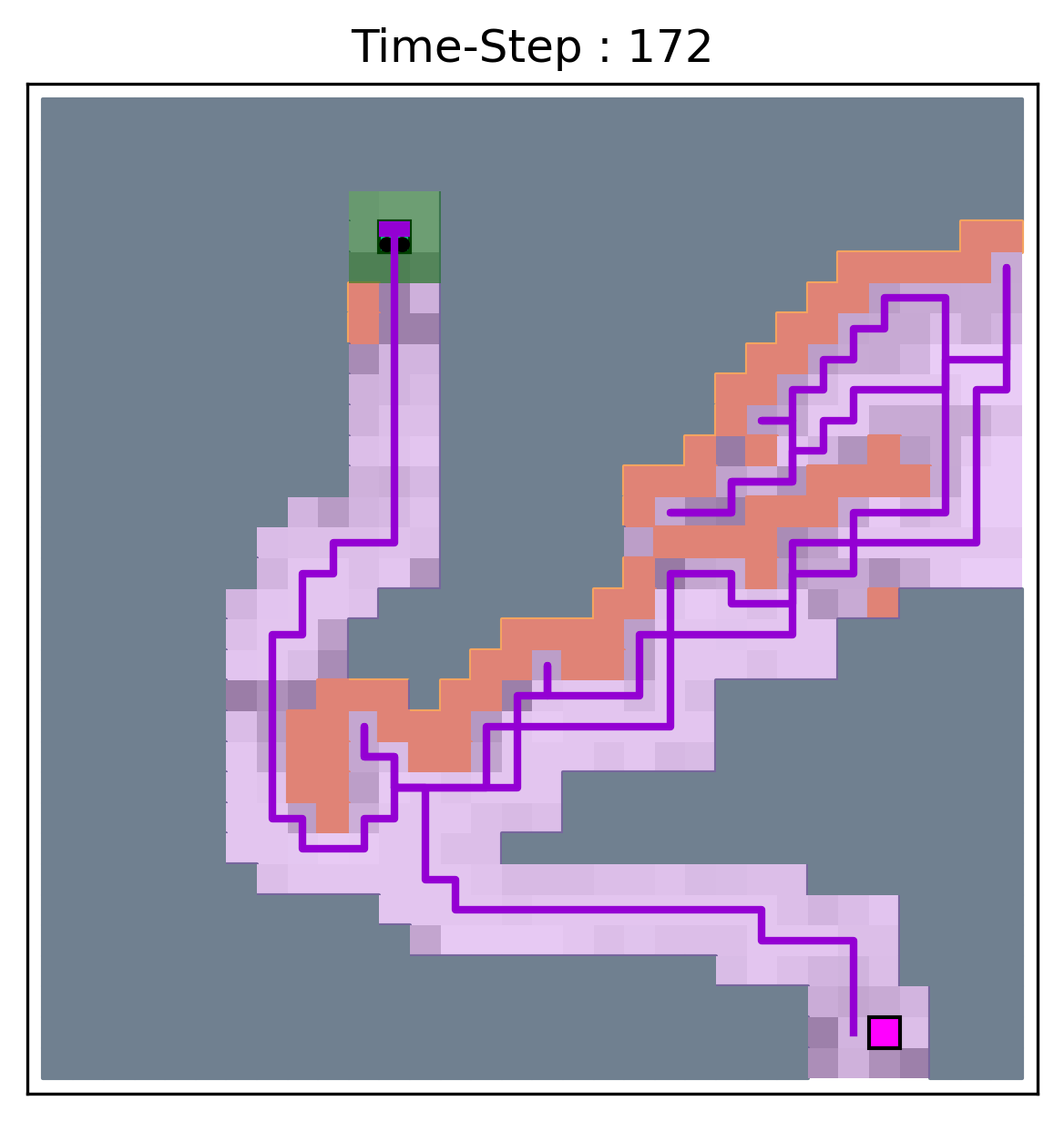}
    \label{Fig:9c}
    }
    \par\hspace{10mm}
    \subfigure{
    \includegraphics[width=0.27\linewidth]{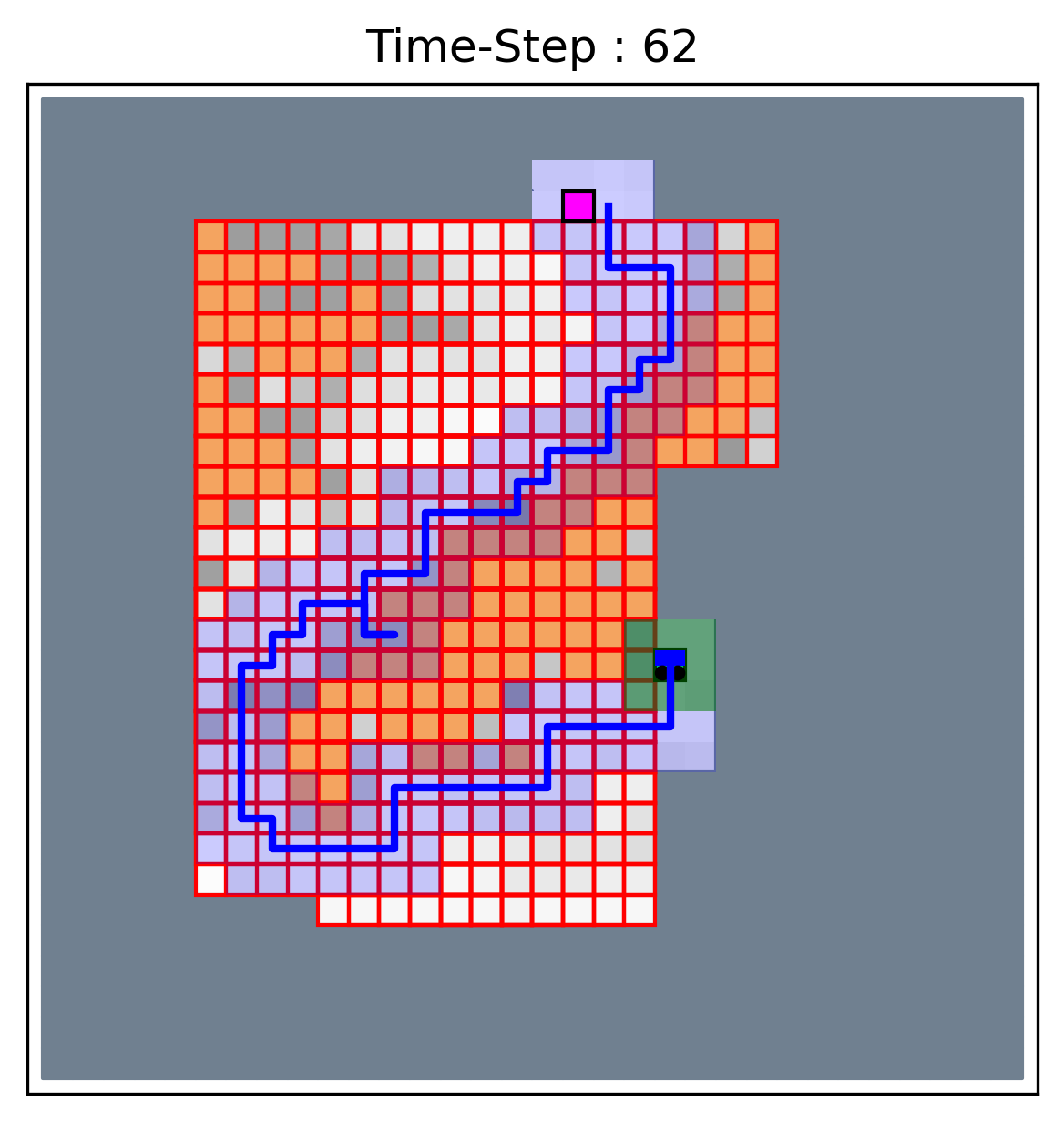}
    \label{Fig:9d}
    }
    \subfigure{
    \includegraphics[width=0.27\linewidth]{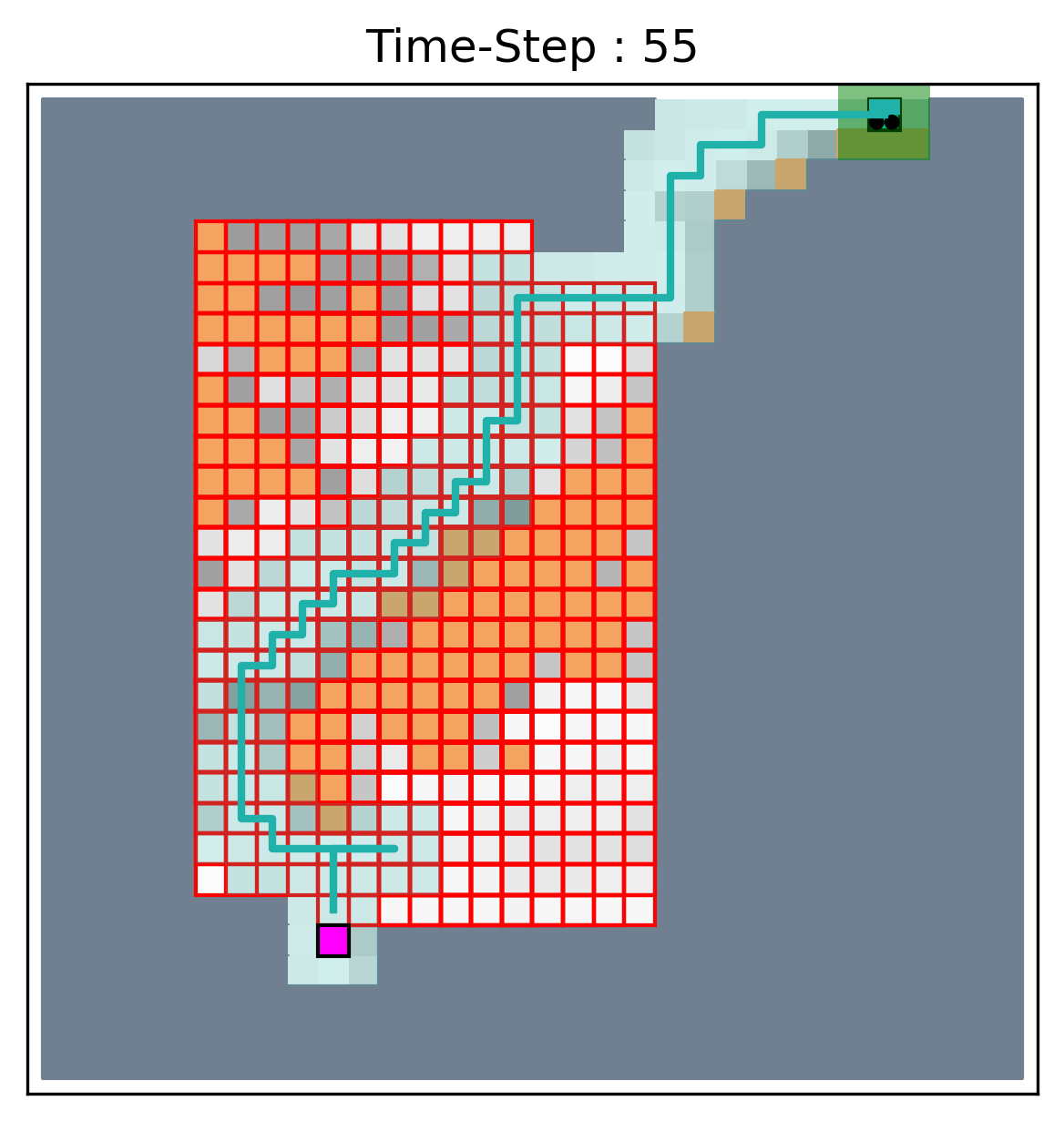}
    \label{Fig:9e}
    }
    \subfigure{
    \includegraphics[width=0.27\linewidth]{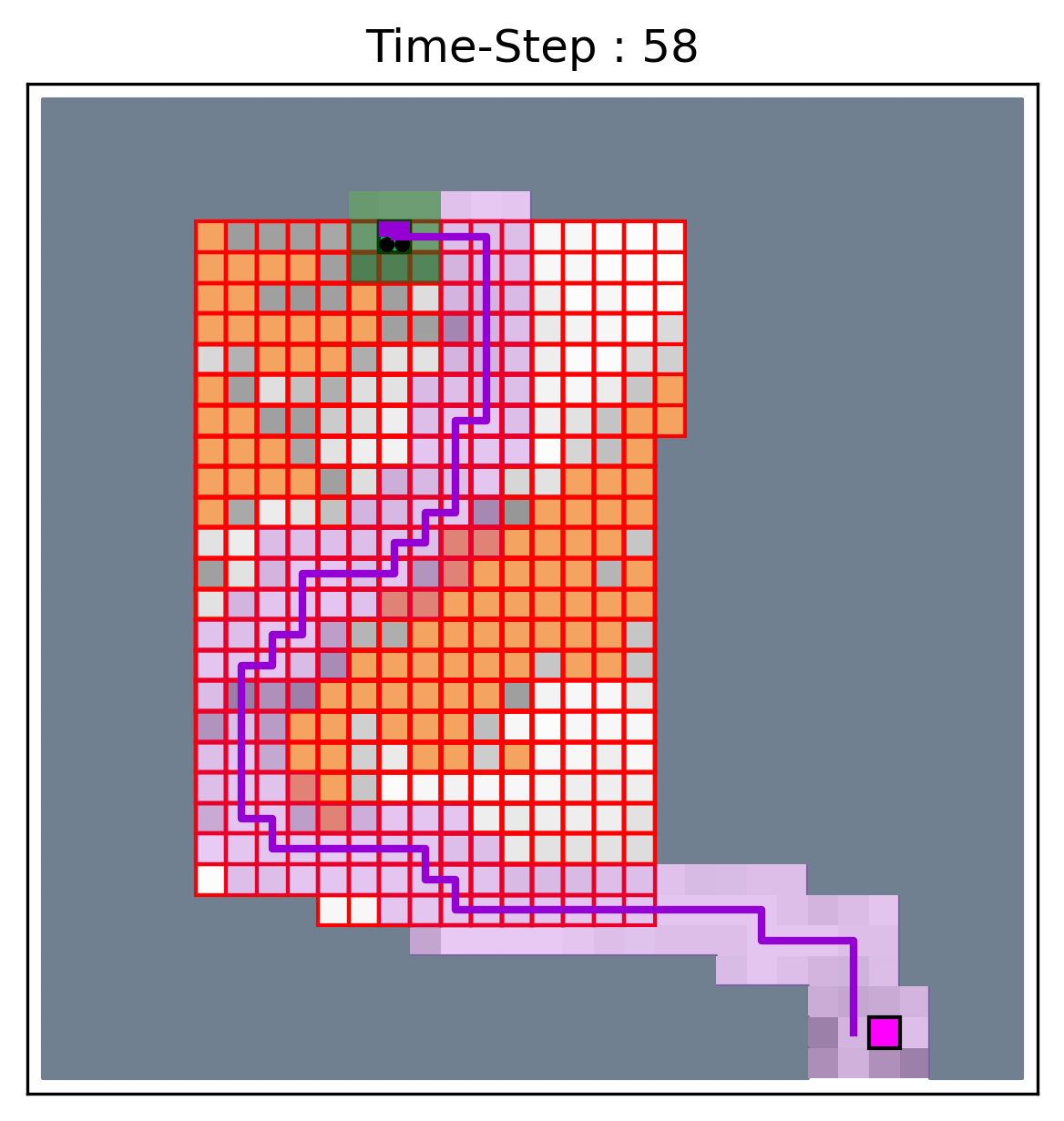}
    \label{Fig:9f}
    }
    \par\hspace{10mm}
    \subfigure{
    \includegraphics[width=0.27\linewidth]{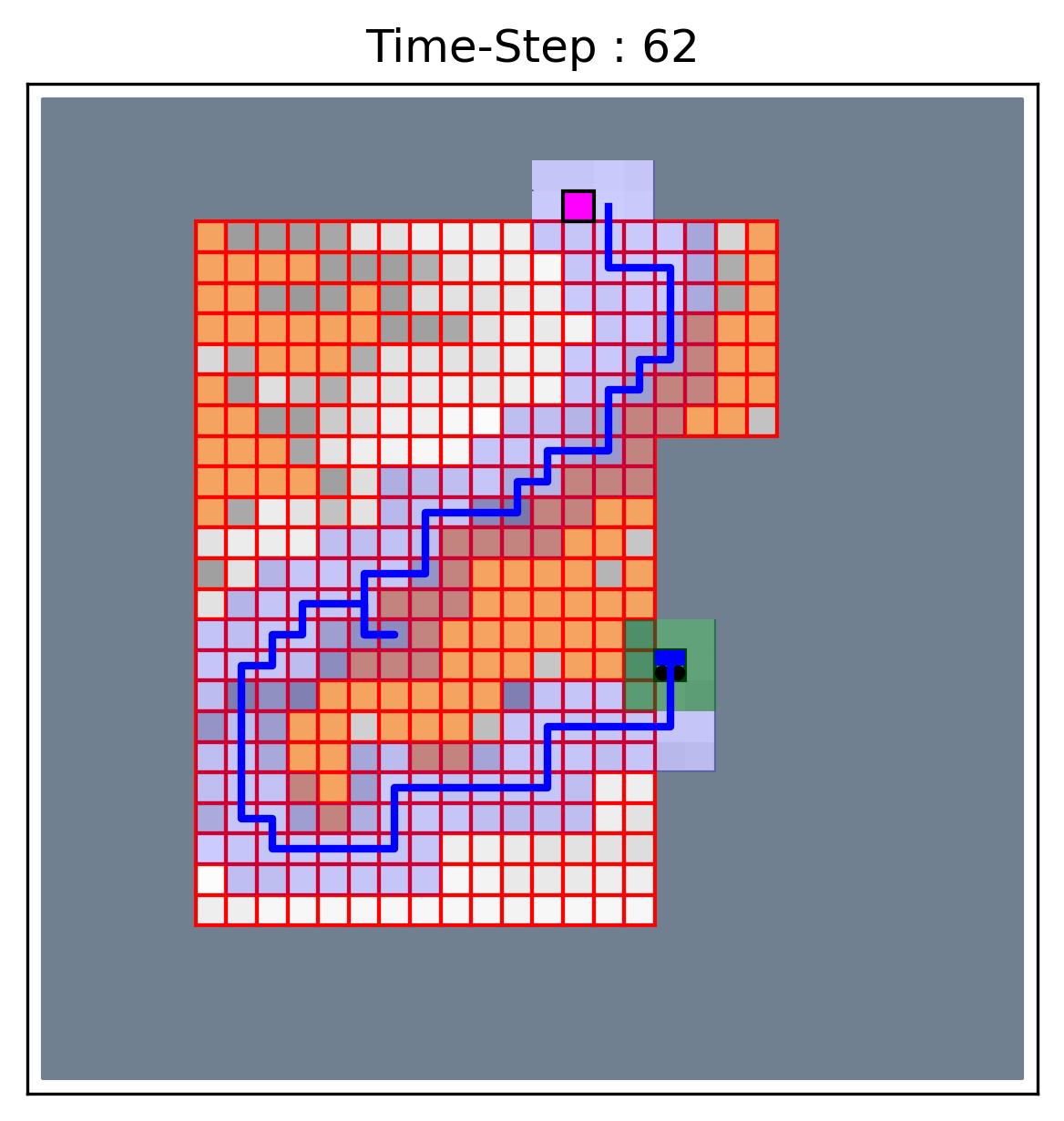}
    \label{Fig:9g}
    }
    \subfigure{
    \includegraphics[width=0.27\linewidth]{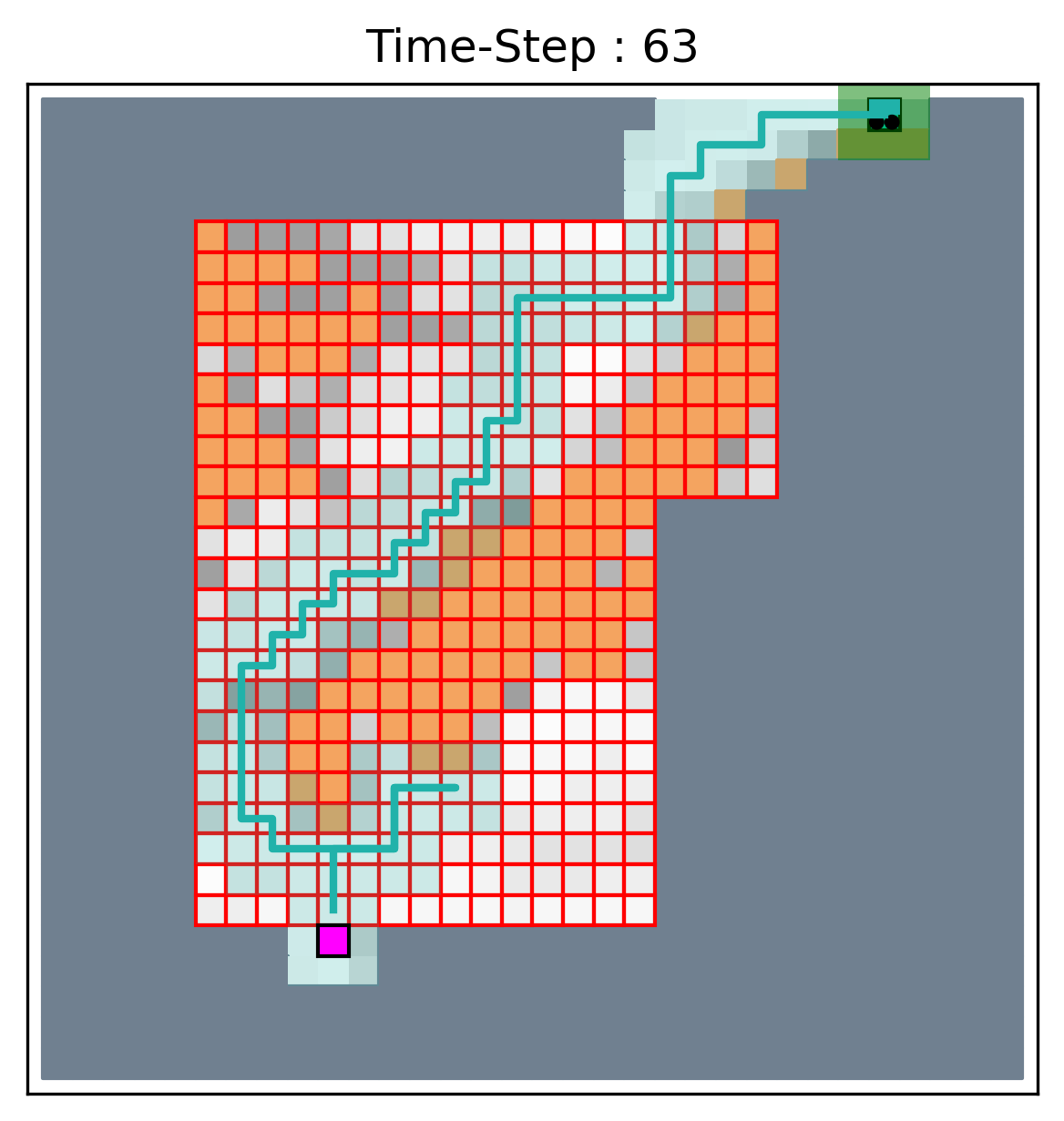}
    \label{Fig:9h}
    }
    \subfigure{
    \includegraphics[width=0.27\linewidth]{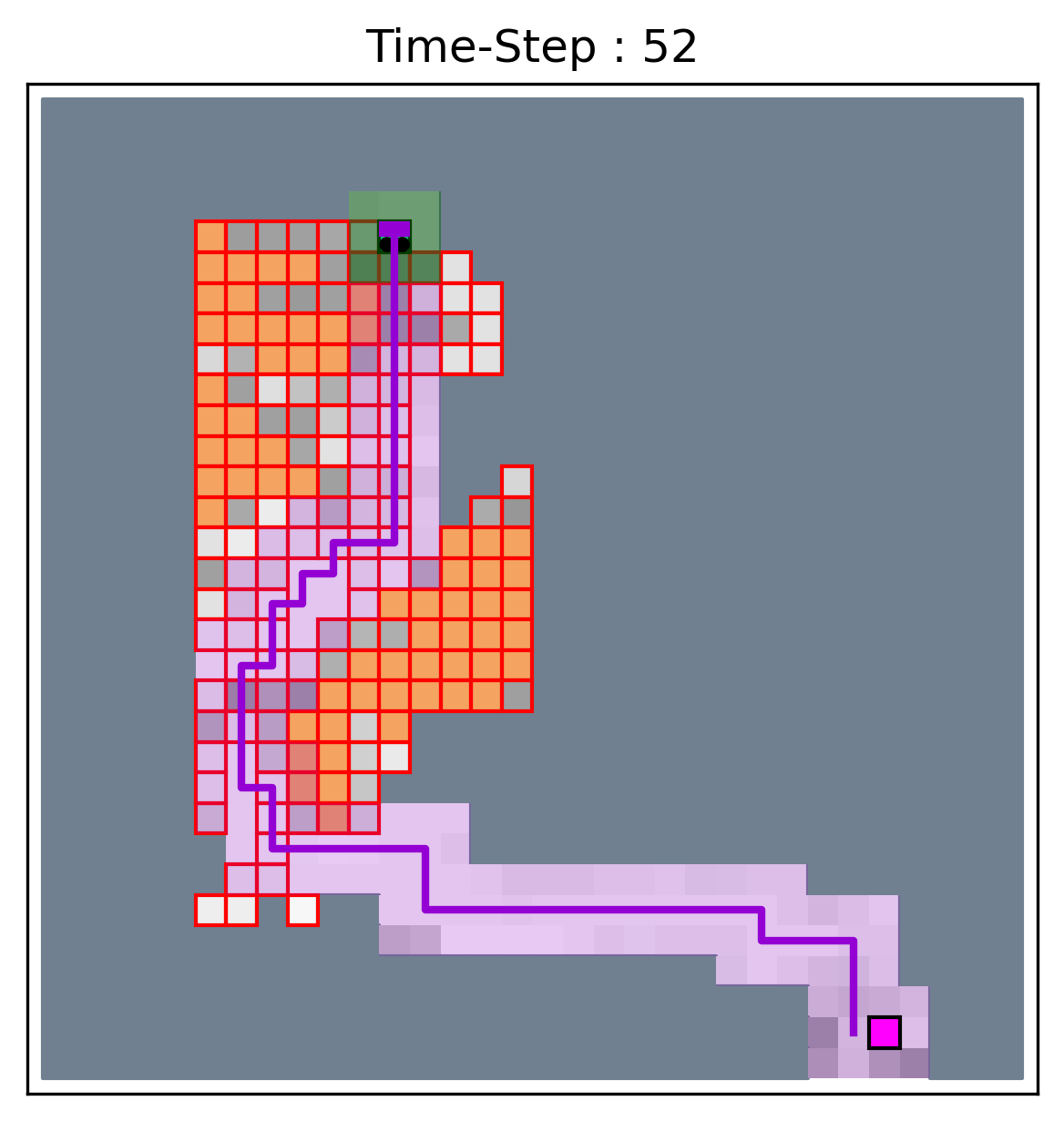}
    \label{Fig:9i}
    }
    \caption{\small Simulation frames of a three-seeker–one-supporter team operating on a $32\times32$ terrain. Each column shows a seeker’s final exploration map at its goal location, with the team of seekers shown in blue, lightseagreen, and dark-violet for the first, second, and third columns, respectively. The regions explored by each seeker are highlighted with its corresponding transparent color shade, while the supporter-transmitted cell locations up to each seeker reaching its final goal are marked with red edges. The magenta cell indicates the seeker’s start location, and the final navigation path is shown as a line in its respective color. The supporter performs default lawn-mower based exploration for the team. The rows represent different communication strategies: uninformed (UI) exploration in the first row, fully informed (FI$_0$) exploration in the second, and the proposed VoI- and MILP$_0$-based method in the third, with maximum allowed bandwidth $B = 27$ for the MILP$_0$ case.}
    
    \begin{tikzpicture}[remember picture, overlay,
    rowtitlenode/.style={rectangle, draw=red!60, fill=red!5, very thick, minimum size=5mm},
    coltitlenode/.style={rectangle, draw=blue!60, fill=blue!5, very thick, minimum size=5mm}]
        \begin{scope}[shift={(0.0,2.5)}]
            \draw[gray, ultra thick, dashed, opacity = 0.5] (-7.5cm, 1.0cm) -- (-7.5cm, 17.0cm); 
            \draw[gray, ultra thick, dashed, opacity = 0.5] (-2.1cm, 1.0cm) -- (-2.1cm,17.0cm); 
            \draw[gray, ultra thick, dashed, opacity = 0.5] (3.25cm, 1.0cm) -- (3.25cm, 17.0cm);
            \draw[gray, ultra thick, dashed, opacity = 0.5] (8.65cm, 1.0cm) -- (8.65cm, 17.0cm);
            \draw[gray, ultra thick, dashed, opacity = 0.5] (-7.5cm, 1.0cm) -- (8.65cm, 1.0cm);
            \draw[gray, ultra thick, dashed, opacity = 0.5] (-7.5cm, 6.25cm) -- (8.65cm, 6.25cm); 
            \draw[gray, ultra thick, dashed, opacity = 0.5] (-7.5cm, 11.70cm) -- (8.65cm, 11.70cm); 
            \draw[gray, ultra thick, dashed, opacity = 0.5] (-7.5cm, 17.0cm) -- (8.65cm, 17.0cm);
            \node[rowtitlenode,rotate=90] at (-8.5cm, 14.25cm) {$\boldsymbol{UI}$};
            \node[rowtitlenode,rotate=90] at (-8.5cm, 9.0cm) {$\boldsymbol{FI}$};
            \node[rowtitlenode,rotate=90] at (-8.5cm, 3.4cm) {$\boldsymbol{MILP}$};
            \node[coltitlenode] at (-4.65cm, 17.5cm) {\textbf{Seeker 1}};
            \node[coltitlenode] at (0.60cm, 17.5cm) {\textbf{Seeker 2}};
            \node[coltitlenode] at (5.85cm, 17.5cm) {\textbf{Seeker 3}};
        \end{scope}
    \end{tikzpicture}
    \label{Figure: 9}
\end{figure*}

The trajectories and the final maps of all three seekers are presented in Figure~\ref{Figure: 8} where the top, middle, and bottom rows correspond to the UI, FI$_1$, and MILP$_1$ methods, respectively. 
Similarly, in Figure~\ref{Figure: 9}, we present the result from the UI, FI$_0$, and MILP$_0$ methods.

It can be observed from Figure~\ref{Figure: 8} and Table~\ref{Table:1} that under the UI framework, \textit{Seeker 2} and \textit{Seeker 3} tend to explore significantly more before reaching their goals compared with the FI$_1$ and MILP$_1$ frameworks.
The total navigation costs for both MILP$_1$ and FI$_1$ are nearly identical, demonstrating that strategic information transfer can achieve similar performance to FI with a fraction of the data communicated.
%

Similar results for each seeker are shown in Figure~\ref{Figure: 9} and Table~\ref{Table:2} for the UI, FI$_0$, and MILP$_0$ methods.
As expected, compared to the utility-based exploration strategy, the lawn-mower strategy results in higher total navigation cost but lower data transmission.
This occurs because utility-based transmission allows the supporter to gather and share relevant observations more quickly, enabling more efficient seeker navigation.

Although Figures~\ref{Figure: 8}–\ref{Figure: 9} show only a slight difference in the number of cells transmitted between the FI and MILP methods, the actual amount of data sent using FI is much higher, as illustrated in Tables~\ref{Table:1}–\ref{Table:2}.
This is because, in FI, the supporter transmits its entire local map observation at each time step throughout the simulation, resulting in a large amount of redundant data being repeatedly sent.
In contrast, under MILP, each cell in the seeker’s exploration map is transmitted exactly once to the seeker over the entire simulation run, avoiding redundant transmissions and reducing the overall data transfer.

\newcolumntype{C}{>{\centering\arraybackslash}p{2cm}}
\begin{table*}
    \centering
    \caption{Simulation data for a team of three-seeker-one-supporter team at baseline and proposed methods and supporter performing utility based exploration on $32\times32$ terrain map environment.}
    \begin{tabular}{|C|C|C|C|C|C|}
        \hline
        \rowcolor{gray!20}
        \multicolumn{6}{|c|}{\rule{0pt}{0.3cm}\textbf{Simulation Data for Single Iteration on Terrain Map Environment (Utility-Based Exploration)}}\\
        \hline
        \rule{0pt}{0.3cm}
        Framework ($\mathbf{F}$) & \cellcolor{white}{\small \textcolor{blue}{Seeker 1 Cost}} & \cellcolor{white}{\small \textcolor{BlueGreen}{Seeker 2 Cost}} & \cellcolor{white}{\small \textcolor{violet}{Seeker 3 Cost}} & \cellcolor{white}{\small \textcolor{black}{Total Cost}} & \cellcolor{white}{\small \textcolor{black}{Supporter Data}} \\
        \hline
        \rule{0pt}{0.3cm}
        UI & 1060.1 & 2599.7 & 2977.8 & 6637.6 & 0 \\
        \hline
        \rule{0pt}{0.3cm}
        FI$_1$ & 917.7 & 426.5 & 634.8 & 1979.0 & 9996 \\
        \hline
        \rule{0pt}{0.3cm}
        MILP$_1$ & 748.6 & 498.3 & 710.4 & 1957.3 & 1041 \\
        \hline
    \end{tabular}
    \label{Table:1}
\end{table*}

\newcolumntype{C}{>{\centering\arraybackslash}p{2cm}}
\begin{table*}
    \centering
    \vspace{-1.25cm}
    \caption{Simulation data for a team of three-seeker-one-supporter team at baseline and proposed methods and supporter performing lawn mower based exploration on $32\times32$ terrain map environment.}
    \begin{tabular}{|C|C|C|C|C|C|}
        \hline
        \rowcolor{gray!20}
        \multicolumn{6}{|c|}{\rule{0pt}{0.3cm}\textbf{Simulation Data for Single Iteration on Terrain Map Environment (Lawn-Mower Exploration)}}\\
        \hline
        \rule{0pt}{0.3cm}
        Framework ($\mathbf{F}$) & \cellcolor{white}{\small \textcolor{blue}{Seeker 1 Cost}} & \cellcolor{white}{\small \textcolor{BlueGreen}{Seeker 2 Cost}} & \cellcolor{white}{\small \textcolor{violet}{Seeker 3 Cost}} & \cellcolor{white}{\small \textcolor{black}{Total Cost}} & \cellcolor{white}{\small \textcolor{black}{Supporter Data}} \\
        \hline
        \rule{0pt}{0.3cm}
        UI & 1060.1 & 2599.7 & 2977.8 & 6637.6 & 0 \\
        \hline
        \rule{0pt}{0.3cm}
        FI$_0$ & 917.7 & 481.3 & 639.0 & 2038.0 & 9114 \\
        \hline
        \rule{0pt}{0.3cm}
        MILP$_0$ & 917.7 & 538.2 & 699.6 & 2155.5 & 925 \\
        \hline
    \end{tabular}
    \label{Table:2}
\end{table*}

\section{Conclusion} \label{Conclusion}


This paper addresses the joint challenge of exploration~and communication for assisting multiple agents.
In the proposed framework, agents collaborate in an unknown environment, where the supporter (helper) performs utility-based exploration and selects map data based on the \textit{Value-of-Information} (VoI) principle to transmit to the seekers.~Bandwidth for each seeker (receiver) is allocated by solving an MILP optimization. 
Simulation results demonstrate improved navigation performance with reduced data transmission.

\bibliographystyle{IEEEtran}
\bibliography{reference_updated}

\end{document}